\documentclass{wscpaperproc}
\usepackage{latexsym}
\usepackage{graphicx}
\usepackage{mathptmx}
\usepackage[T1]{fontenc}

\usepackage{amsmath}
\usepackage{amsfonts}
\usepackage{amssymb}
\usepackage{amsbsy}
\usepackage{amsthm}
\usepackage{bm}
\usepackage{multirow, booktabs}
\usepackage{algorithm}
\usepackage{algorithmic}
\usepackage{enumitem}
\usepackage{epstopdf}
\usepackage{graphicx}
\usepackage{subcaption}
\usepackage[pdftex,colorlinks=true,urlcolor=blue,citecolor=blue,anchorcolor=black,linkcolor=black]{hyperref}

\newtheoremstyle{wsc}
{3pt}
{3pt}
{}
{}
{\bf}
{}
{.5em}
{}

\theoremstyle{wsc}

\theoremstyle{remark}

\newcommand{\var}{\mbox{var}}

\DeclareMathOperator*{\E}{E}
\DeclareMathOperator*{\argmax}{arg\,max} 

\def\cN{\mathcal{N}}

\newcommand{\dif}{\,\mathrm{d}}

\graphicspath{{figure2/}}

\begin{document}

%
%

\pagestyle{fancyplain}

\thispagestyle{plain}
\firstPageHead{}

\chead{\fancyplain{}{\itshape Cheng, Kang, Wang, Liu}}

\rhead{}
\cfoot{}
\renewcommand{\headrulewidth}{0pt} 

\makeatletter
\let\@internalcite\cite
\def\cite{\def\@citeseppen{-1000}%
    \def\@cite##1##2{(##1\if@tempswa , ##2\fi)}%
    \def\citeauthoryear##1##2##3{##1 ##3}\@internalcite}
\def\citeNP{\def\@citeseppen{-1000}%
    \def\@cite##1##2{##1\if@tempswa , ##2\fi}%
    \def\citeauthoryear##1##2##3{##1 ##3}\@internalcite}
\def\citeN{\def\@citeseppen{-1000}%
    \def\@cite##1##2{##1\if@tempswa, ##2)\else{}\fi}%
    \def\citeauthoryear##1##2##3{##1 (##3)}\@citedata}
\def\citeA{\def\@citeseppen{-1000}%
    \def\@cite##1##2{(##1\if@tempswa , ##2\fi)}%
    \def\citeauthoryear##1##2##3{##1}\@internalcite}
\def\citeANP{\def\@citeseppen{-1000}%
    \def\@cite##1##2{##1\if@tempswa , ##2\fi}%
    \def\citeauthoryear##1##2##3{##1}\@internalcite}
\def\shortcite{\def\@citeseppen{-1000}%
    \def\@cite##1##2{(##1\if@tempswa , ##2\fi)}%
    \def\citeauthoryear##1##2##3{##2 ##3}\@internalcite}
\def\shortciteNP{\def\@citeseppen{-1000}%
    \def\@cite##1##2{##1\if@tempswa , ##2\fi}%
    \def\citeauthoryear##1##2##3{##2 ##3}\@internalcite}
\def\shortciteN{\def\@citeseppen{-1000}%
    \def\@cite##1##2{##1\if@tempswa, ##2\else{}\fi}%
    \def\citeauthoryear##1##2##3{##2 (##3)}\@citedata}
\def\shortciteA{\def\@citeseppen{-1000}%
    \def\@cite##1##2{(##1\if@tempswa , ##2\fi)}%
    \def\citeauthoryear##1##2##3{##2}\@internalcite}
\def\shortciteANP{\def\@citeseppen{-1000}%
    \def\@cite##1##2{##1\if@tempswa , ##2\fi}%
    \def\citeauthoryear##1##2##3{##2}\@internalcite}
\def\citeyear{\def\@citeseppen{-1000}%
    \def\@cite##1##2{(##1\if@tempswa , ##2\fi)}%
    \def\citeauthoryear##1##2##3{##3}\@citedata}
\def\citeyearNP{\def\@citeseppen{-1000}%
    \def\@cite##1##2{##1\if@tempswa , ##2\fi}%
    \def\citeauthoryear##1##2##3{##3}\@citedata}
%
%
%
\def\@citedata{%
    \@ifnextchar [{\@tempswatrue\@citedatax}%
                  {\@tempswafalse\@citedatax[]}%
}

\def\@citedatax[#1]#2{%
\if@filesw\immediate\write\@auxout{\string\citation{#2}}\fi%
  \def\@citea{}\@cite{\@for\@citeb:=#2\do%
    {\@citea\def\@citea{, }\@ifundefined
       {b@\@citeb}{{\bf ?}%
       \@warning{Citation `\@citeb' on page \thepage \space undefined}}%
{\csname b@\@citeb\endcsname}}}{#1}}%

%
\def\@citex[#1]#2{%
\if@filesw\immediate\write\@auxout{\string\citation{#2}}\fi%
  \def\@citea{}\@cite{\@for\@citeb:=#2\do%
    {\@citea\def\@citea{; }\@ifundefined
       {b@\@citeb}{{\bf ?}%
       \@warning{Citation `\@citeb' on page \thepage \space undefined}}%
{\csname b@\@citeb\endcsname}}}{#1}}%

%
\def\@biblabel#1{}
\makeatother



\newdimen\bibindent
\bibindent=0.0em
\def\thebibliography#1{\section*{\refname}\list
   {}{\settowidth\labelwidth{[#1]}
   \leftmargin\parindent
   \itemindent -\parindent
   \listparindent \itemindent
   \itemsep 0pt
   \parsep 0pt}
   \def\newblock{}
   \sloppy
   \sfcode`\.=1000\relax}


\setlength{\baselineskip}{12.7pt}

\title{Active Learning for Manifold Gaussian Process Regression}

\author{\begin{center}Yuanxing Cheng\textsuperscript{1},  Lulu Kang\textsuperscript{2}, Yiwei Wang\textsuperscript{3}, and Chun Liu\textsuperscript{1}\\
[11pt]
\textsuperscript{1}Dept.~of Applied Mathematics, Illinois Institute of Technology, Chicago, IL, USA\\
\textsuperscript{1}Dept.~of Mathematics and Statistics, University of Massachusetts, Amherst, MA, USA\\
\textsuperscript{3}Dept.~of Mathematics, University of California, Riverside, CA, USA
\end{center}
}

\maketitle

\vspace{-12pt}

\section*{ABSTRACT}
This paper introduces an active learning framework for manifold Gaussian Process (GP) regression, combining manifold learning with strategic data selection to improve accuracy in high-dimensional spaces. Our method jointly optimizes a neural network for dimensionality reduction and a Gaussian process regressor in the latent space, supervised by an active learning criterion that minimizes global prediction error. Experiments on synthetic data demonstrate superior performance over randomly sequential learning. The framework efficiently handles complex, discontinuous functions while preserving computational tractability, offering practical value for scientific and engineering applications. Future work will focus on scalability and uncertainty-aware manifold learning.

\section{INTRODUCTION}
\label{sec:intro}
Gaussian Process Regression (GPR) \shortcite{rasmussen2006gaussian} is a powerful non-parametric framework widely used in machine learning, statistics, and scientific computing for regression and uncertainty quantification tasks. However, standard GPR methods face significant challenges as the input dimension grows, suffering from the \emph{curse of dimensionality}: computational scalability deteriorates, and the ability to capture the underlying data structure diminishes as data becomes increasingly sparse in high-dimensional spaces \cite{beyer_when_1999}.

To address these limitations, \emph{manifold Gaussian Processes} (mGPs) have emerged as a promising alternative. By leveraging manifold learning techniques, mGPs project high-dimensional data onto a lower-dimensional latent space while preserving its intrinsic geometry \shortcite{meila_manifold_2024}. This approach not only mitigates the challenges of high-dimensional inference but also enhances predictive performance by exploiting the data's underlying structure. Recent advances include graph-based distance metrics (e.g., Isomap \shortcite{tenenbaum_global_2000}) combined with Mat\'{e}rn Gaussian processes \shortcite{borovitskiy_matern_2023,fichera_implicit_2024}, as well as wrapped Gaussian processes for manifold-valued data \shortcite{mallasto_wrapped_2018}.

Despite these innovations, the performance of GPR—and by extension, mGPs—heavily depends on the quality and representativeness of the training data. In many real-world applications, acquiring labeled data is expensive or time-consuming, necessitating efficient data acquisition strategies. Active learning for GPR has been extensively studied due to the model's inherent ability to quantify predictive uncertainty, making it a natural fit for sequential data selection. Early work by \cite{mackay1992information} laid the theoretical foundation for Bayesian active learning, later adapted to GPR by \shortciteN{seo2000gaussian}, demonstrating superior convergence compared to passive sampling. Subsequent developments include uncertainty sampling (prioritizing high-variance points) and information-theoretic approaches, such as mutual information maximization for sensor placement \shortcite{krause2008near} and Bayesian Active Learning by Disagreement (BALD) for parameter-aware selection \shortcite{houlsby2011bayesian}. These methods have been extended to deep GPR models \shortcite{ma2019variational} and related tasks like uncertainty quantification and inverse problems \shortcite{Chen03072021,heo2025active}.

In practice, active learning strategies must balance \emph{exploration} (reducing global uncertainty) and \emph{exploitation} (optimizing a downstream objective). This trade-off is well-studied in Bayesian optimization (BO) \cite{frazier2018bayesian}, where acquisition functions like Expected Improvement (EI) and Upper Confidence Bound (GP-UCB) \shortcite{srinivas2010gaussian} align naturally with active learning objectives. For instance, GP-UCB provides theoretical regret bounds, making it suitable for bandit-style optimization tasks. Cost-sensitive active learning methods further refine this balance by incorporating labeling costs into the acquisition function \shortcite{kapoor2007active}, which is critical in domains like medical diagnosis. Recent work has also focused on robustness: \shortciteN{pmlr-v84-martinez-cantin18a} combined robust regression (GPs with Student-t likelihood) with outlier diagnostics to improve reliability.

In the context of mGPs, active learning can efficiently refine the model by targeting regions of high uncertainty or high potential improvement. For example, \shortciteN{Sauer02012023} and \shortciteN{binois_replication_2019} adopted the \emph{integrated mean-squared error} (IMSE) criterion—originally proposed by \shortciteN{cohn_active_1994,cohn_neural_1996}—to minimize prediction error. \shortciteN{doi:10.1137/22M1529907} further integrated manifold learning with GP-UCB for optimization on manifolds. However, \emph{active learning tailored to mGPs remains underexplored}, particularly for regression tasks where latent space geometry and high-dimensional inputs interact dynamically.

This paper proposes a novel active learning framework for manifold Gaussian Process regression, bridging the gap between efficient data acquisition and dimensionality reduction. Our key contributions include:
\begin{itemize}
\item a unified framework integrating manifold learning with IMSE-based active learning for mGPs,
\item enhanced efficiency through strategic point selection in the latent space, and
\item scalability for high-dimensional data by operating on intrinsic low-dimensional manifolds.
\end{itemize}
We validate our approach through synthetic experiments, demonstrating improvements over passive sampling and standard active learning methods.

The paper is organized as follows: Section~\ref{sec:background} reviews GPR fundamentals. Section~\ref{sec:ALMGP} introduces our active learning method with active learning Cohn (ALC) acquisition function for mGP regression. Section~\ref{sec:examples} illustrates the framework with three case studies, and Section~\ref{sec:conclusion} discusses future directions.

\section{Background on GPR}\label{sec:background}

The Gaussian Process Regression (GPR) model assumes the following probabilistic representation for the response $y(\bm x)$:
\[
y(\bm x)=\mu(\bm x)+Z(\bm x)+\epsilon,
\]
where
\[
Z(\bm x)\sim GP(0, \tau^2 k(\cdot, \cdot)),\quad\textrm{ and } \epsilon \overset{\underset{\mathrm{iid}}{}}{\sim} N(0,\sigma^2).
\]
Denote $\Omega \subset \mathbb{R}^p$ as the input domain and output $y(\bm x)\in \mathbb{R}^1$. 
Here, $k(\cdot,\cdot): \Omega\times \Omega \to \mathbb{R}_+$ is the correlation function of the process $Z(\bm x)$.
Common choices include the Mat\'{e}rn and Gaussian correlation functions (or SE-ARD kernels):
\[
\textrm{Mat\'{e}rn: }\prod_{l=1}^p\frac{1}{\Gamma(\nu)2^{\nu-1}}\left(\frac{2\sqrt{\nu}|x_{il}-x_{jl}|}{\theta_l}\right)^{\nu}K_{\nu}\left(\frac{2\sqrt{\nu}|x_{il}-x_{jl}|}{\theta_l}\right), \quad 
\textrm{Gaussian: }\exp\left\{-\sum_{l=1}^p\frac{(x_{il}-x_{jl})^2}{\theta_l}\right\},
\]
where $K_{\nu}$ is the modified Bessel function of the second kind and  $\bm \theta\in\mathbb{R}^p_{+}$ are length-scale parameters.
The mean function $\mu(\bm x)$ determines the model class:
\begin{itemize}
\item \emph{Ordinary kriging}: $\mu(\bm x)$ is an unknown constant; 
\item \emph{Universal kriging}: $\mu(\bm x)=\beta_0+\bm \beta^\top \bm x$  (linear in inputs); 
\item \emph{General case}: $\mu(\bm x)$ is an unknown combination of basis functions, selectable via stepwise forward regression \shortcite{joseph2008blind,joseph2011regression} or shrinkage methods \shortcite{hung2011penalized,kang2024energetic}. 
\end{itemize}
For simplicity, we assume $\mu(\bm x)=0$ throughout this paper. The proposed methods can be extended to non-zero mean cases through response centering or trend removal in preprocessing.

Given observed data $\{\bm x_i, y_i\}_{i=1}^n$, we need to estimate the unknown parameters that the length-scale parameters $\bm \theta$, the variance of $Z(\cdot)$ $\tau^2$, and the variance of the noise $\sigma^2$.
To estimate $\sigma^2$, if the training data contain replications at some if not all design points, then $\sigma^2$ can be estimated in advance using pooled sample variance. 
The response vector $\bm y$ follows the multivariate normal distribution $\bm y\sim \cN({\bf 0}, \tau^2\bm K_n+\sigma^2\bm I_n)$, where $\bm K_n$ is the $n\times n$ correlation matrix with $\bm K_{n,ij}=k(\bm x_i, \bm x_j)$, and $\bm I_n$ is the $n$-dim diagonal matrix. 

The parameters $(\bm\theta, \tau^2, \sigma^2)$ can be estimated via maximum likelihood method, and the $-2\log$-likelihood of the data is
\begin{align}
\label{eq:likelihood}
Q\triangleq-2\log L=n\log\tau^2+\log \det\left(\bm K_n+\rho\bm I_n\right)+\frac{1}{\tau^2}\left(\bm y^\top\left(\bm K_n+\rho \bm I_n\right)^{-1}\bm y\right).
\end{align}
Here $\rho=\sigma^2/\tau^2$ is the noise-to-signal ratio.
If $\rho=0$, the GPR becomes an interpolation model which is usually used to model deterministic computer simulation outputs.
If the data does not contain replications at any design point and the noise variance $\sigma^2$ is not zero, we can consider $\rho$ as an unknown parameter and estimate it by maximizing the log-likelihood. 
Given the other parameter values, the MLE of $\tau^2$ is
\[
\hat{\tau}^2=\frac{1}{n}\bm y^\top (\bm K_n+\rho\bm I_n)^{-1}\bm y
\]
obtained by solving $\partial Q/\partial \tau^2=0$.
Replace $\tau^2$ by $\hat{\tau}^2$ in $Q$ and minimize $Q$ with respect to $(\bm \theta, \rho)$ if $\sigma^2$ cannot be estimated from the pooled variance from replicated observations. 

The predictive distribution of $y$ at any new query point $\bm x$ is also normally distributed with 
\begin{align}
\label{eq:pred-mean}
\E(y(\bm x)|\bm y, \tau^2, \rho, \bm \theta)&=\bm k_n(\bm x)'(\bm K_n+\hat{\rho} \bm I_n)^{-1}\bm y,\\
\label{eq:pred-var}
s_{n}^2(\bm x)\triangleq\var(y(\bm x)|\bm y, \tau^2, \rho, \bm \theta)&=\hat{\tau}^2\left[1-\bm k(\bm x)'(\bm K_n+\hat{\rho}\bm I_n)^{-1}\bm k(\bm x)\right]+\hat{\sigma}^2,
\end{align}
where $\bm k_n(\bm x)$ is the correlation between $\bm x$ and $\bm x_i$, i.e., $\bm k_n(\bm x)=[k(\bm x,\bm x_1), \ldots, k(\bm x, \bm x_n)]^\top$. 
The GPR predictor is the conditional mean $\hat{y}(\bm x)=\bm k_n(\bm x)'(\bm K_n+\hat{\rho} \bm I_n)^{-1}\bm y$. 
It is also the Best Linear Unbiased Predictor (BLUP) \shortcite{santner2003design} for $y(\bm x)$ under the GP assumption. 
We can construct the predictive confidence interval for $y(\bm x)$ from \eqref{eq:pred-var}.

\section{Active Learning for Manifold Gaussian Process Regression}\label{sec:ALMGP}

\subsection{Manifold Gaussian Process Regression}\label{subsec:mGP}

The \emph{Manifold Gaussian Process} (mGP) \cite{calandra_manifold_2016} addresses key limitations of standard Gaussian Process Regression (GPR) by jointly learning a data transformation and a GPR model. Standard GPs rely on covariance functions (e.g., squared exponential) that encode smoothness assumptions, which may be inadequate for modeling discontinuous or complex functions (e.g., robotics tasks with contact dynamics). The mGP framework decomposes the regression task into:
\[F=G\circ M,\]
where $M: \mathcal{X}\to \mathcal{H}$ is a deterministic mapping from the input space $\mathcal{X}$ to a latent feature space $\mathcal{H}\subset \mathbb{R}^Q$, and $G:\mathcal{H}\to \mathcal{Y}$ is a GPR model on $\mathcal{H}$. 
Here $\mathcal{Y}\subset \mathbb{R}$ is the space of the response. 
For the map $M$, we employ multilayer neural networks. 
In each layer, the transformation is defined as $\bm Z_i=T_i(\bm X)=\sigma_M(\bm W_i \bm Z_{i-1}+\bm B_i)$, where $\sigma_M$ is the activation function and $\bm W_i$ and $\bm B_i$ are the weight matrix and bias vector for the $i$-th layer. 
Consequently, $M(\bm X)=T_l\circ T_{l-1}\circ\cdots\circ T_1(\bm Z_0)$ with $\bm Z_0=\bm X$, and the collection of these hyperparameters is denoted by $\bm \theta_M$.

The mGP defines a valid GP over $\mathcal{X}$ with the covariance function:
\[\tau^2\tilde{k}(\bm x_1,\bm x_2)=\tau^2k(M(\bm x_1), M(\bm x_2)),\]
where $k$ is the standard kernel (e.g., SE-ARD or neural network kernel) with hyperparameters $\bm \theta_G$. 
This formulation allows the model to adaptively learn feature representations that align with the regression objective, overcoming the limitations of unsupervised transformations (e.g., PCA or random embeddings) that may not preserve the task-relevant structure.

The mGP optimizes the parameters $\bm \theta_{\text{mGP}}=[\bm \theta_M, \bm \theta_G]$ jointly by minimizing the Negative Log Marginal Likelihood (NLML):
\[
\text{NLML}(\bm \theta_{\text{mGP}})=n\log \tau^2+ \log\det(\tilde{\bm K}_n+\rho\bm I_n)+ \frac{1}{\tau^2}\bm y^\top (\tilde{\bm K}_n+\rho \bm I_n)^{-1}\bm y, 
\]
where $\tilde{\bm K}_n$ is  the kernel matrix evaluated on the transformed inputs $M(\bm X)$. 
Gradients are computed via backpropagation through $M$ (if $M$ is a neural network) and the kernel hyperparameters $\bm \theta_G$. 
The predictive distribution at a test point $\bm x_{n+1}$ is a normal distribution with mean and variance specified by \eqref{eq:pred-mean} and 
\eqref{eq:pred-var} except $\bm k_n$ is replaced by $\tilde{\bm k}_n$ and $\bm K_n$ by $\tilde{\bm K}_n$. 

One can see that the Deep Gaussian Process (or DGP) \shortcite{pmlr-v31-damianou13a} is closely connected with mGP since both involve latent space transformations before applying GPR.
However, they are distinct extensions of standard GPR because DGPs learn this transformation hierarchically through layers of GPs, while mGPs often assume a geometric prior and embed the data accordingly.
The mGP framework offers several key advantages over traditional GPs and unsupervised feature learning methods. 
By jointly optimizing the feature mapping $M$ and GPR $G$ under a supervised objective, it learns task-aware representations that can handle discontinuities (e.g., step functions) and multi-scale patterns more effectively than standard kernels like SE-ARD or neural network covariances. 
The flexibility of parameterizing $M$ as a neural network (e.g., $[1-6-2]$ layers) allows it to unwrap complex input geometries, while retaining the GP's probabilistic uncertainty quantification—critical for robotics and control applications.
However, this approach also introduces challenges: the joint optimization $\bm \theta_{\text{mGP}}=[\bm \theta_M, \bm \theta_G]$ is non-convex and may converge to suboptimal local minima, particularly with high-dimensional $\bm \theta_M$.
Additionally, sparse data around discontinuities can lead to overfitting in the deterministic mapping $M$, causing misaligned feature representations. 
While probabilistic extensions (e.g., Bayesian neural networks for $M$) could mitigate this, they often sacrifice tractability. 
Despite these limitations, the mGP's empirical performance on benchmarks in \shortciteN{calandra_manifold_2016} demonstrates its superiority in scenarios where input-space geometry violates standard GP assumptions.

\subsection{Active Learning for mGP}\label{subsec:AL}

The Active Learning Cohn (ALC) criterion \shortcite{cohn_active_1994,seo_gaussian_2000} is a foundational information-theoretic strategy for sequential experimental design in Gaussian Process Regression (GPR). 
It selects new data points $\bm x_{n+1}$ that maximize the model's average predictive uncertainty over the input space. Formally, this is achieved by minimizing the \emph{integrated mean squared error (IMSE)} across the input space \shortcite{cohn_neural_1996}, which corresponds to reducing the average posterior variance. Neglecting the constant noise variance, the IMSE is defined as:
\begin{align*}
\Breve{s}_{n+1}^2(\bm x|\bm x_{n+1})&\triangleq\hat{\tau}^2\left[1-\tilde{\bm k}_{n+1}(\bm x)^\top (\tilde{\bm K}_{n+1}+\rho \bm I_{n+1})^{-1}\tilde{\bm k}_{n+1}(\bm x)\right], \\
\text{IMSE}(\bm x_{n+1})&\triangleq \int_{\mathcal{X}} \Breve{s}_{n+1}^2(\bm x|\bm x_{n+1})\dif \bm x,
\end{align*}
where $\bm x_{n+1}$ is selected from a finite candidate pool $\mathcal X_{\text{cand}}\subset \mathcal X$, typically constructed using a space-filling design. The term $\tilde{\bm k}_{n+1}(\bm x)=[\tilde{k}(\bm x,\bm x_1), \ldots, \tilde{k}(\bm x, \bm x_{n+1})]^\top$ and $\tilde{\bm K}_{n+1}$ denote the kernel vector and $(n+1)\times(n+1)$ correlation matrix evaluated in the learned latent space $\mathcal{H}=M(\mathcal{X})$ using the current mGP parameters.

Minimizing the IMSE is equivalent to maximizing the expected reduction in posterior variance, leading to the ALC acquisition function: 
\[
\text{ALC}(\bm x_{n+1})=\int_{\mathcal{X}} s_n^2(\bm x) - \Breve{s}_{n+1}^2(\bm x|\bm x_{n+1})\dif \bm x \propto \int_{\mathcal{X}} \hat{\tau}^2 \tilde{\bm k}_{n+1}(\bm x)^\top (\tilde{\bm K}_{n+1}+\rho \bm I_{n+1})^{-1}\tilde{\bm k}_{n+1}(\bm x) \dif \bm x,
\]
where $s_{n}^2(\bm x)$ is computed via \eqref{eq:pred-var} except that $k$ function is replaced by $\tilde{k}$, and thus $\bm k_n(\bm x)$ by $\tilde{\bm k}_n(\bm x)$ and $\bm K_n$ by $\tilde{\bm K}_n$. 
To approximate the integration numerically, we introduce a fixed reference set $\mathcal{X}_{\text{ref}}$ of size $m$ sampled from $\mathcal{X}$ using Latin Hypercube Design. The ALC objective becomes:
\[
\text{ALC}(\bm x_{n+1})\propto \sum_{\bm x\in \mathcal{X}_{\text{ref}}} \hat{\tau}^2\tilde{\bm k}_{n+1}(\bm x)^\top (\tilde{\bm K}_{n+1}+\rho \bm I_{n+1})^{-1}\tilde{\bm k}_{n+1}(\bm x). 
\]

In contrast to \shortciteN{Sauer02012023}, which employs DGPs, our method integrates ALC with mGPs. At each active learning iteration, the most informative training point is selected via: 
\[\bm x_{n+1}^* = \argmax_{\bm x \in \mathcal{X}_{\text{cand}}} \text{ALC}(\bm x),\]
and subsequently removed from the candidate pool $\mathcal X_{\text{cand}}$. This selection procedure iteratively reduces overall model uncertainty, leveraging the mGP's ability to capture complex geometric structures in latent space.
To enhance computational efficiency, we adopt two strategies. First, we support batch selection by choosing $B$ design points per iteration, indexed by round $r$. Second, instead of evaluating ALC on the full candidate set, we incorporate a pre-screening step: among the candidate set $\mathcal{X}_{\text{cand}}$, we identify the top $K > B$ points with the highest predictive variance, denoted $\mathcal{X}^*_{r}$:
\[\mathcal{X}_{\text{cand}}^{(r)} = \left\{ \bm x_{1}^{(r)},\bm x_{2}^{(r)},\dots,\bm x_{K}^{(r)} \right\},\]
and compute ALC only on $\mathcal{X}_{\text{cand}}^{(r)}$. 
We then select the $B$ points with the highest ALC scores. 
The complete procedure is outlined in Algorithm~\ref{alg:ALMGP}.

\begin{algorithm}[ht]
\caption{Active Learning for Manifold Gaussian Process Regression (ALmGP)}\label{alg:ALMGP}
\begin{algorithmic}
\STATE \textbf{Input:} 
Initial training dataset $\{\bm x_i, y_i\}_{i=1}^{n_0}$ with small sample size $n_0$; 
reference set $\mathcal{X}_{\text{ref}} \subset \mathcal{X}$ and candidate set $\mathcal{X}_{\text{cand}} \subset \mathcal{X}$; 
tuning parameters including screening size $K$, batch size $B$, convergence threshold \texttt{Tol}, 
maximum total sample size $N_{\max}$, 
and initial values and optimization settings for fitting the mGP.

\STATE \textbf{Output:} Trained mGP model and estimated parameters after active data acquisition.

\STATE \textbf{Step 0:} Fit the mGP model using the training set $\{\bm x_i, y_i\}_{i=1}^{n_0}$ by minimizing the negative log marginal likelihood (NLML) with respect to $\bm \theta_{\text{mGP}}$ using the L-BFGS algorithm \cite{nocedal_numerical_2006}.

\FOR{$r = 1, 2, \ldots, \lfloor N_{\max} / B \rfloor$}
    \STATE \textbf{Step 1:} Compute posterior predictive variance for all points in $\mathcal{X}_{\text{cand}}$, and select the top $K$ points with highest variance. Denote this screened set as $\mathcal{X}_{\text{cand}}^{(r)}$.
    
    \STATE \textbf{Step 2:} For each $\bm x \in \mathcal{X}_{\text{cand}}^{(r)}$, compute the ALC acquisition score $\text{ALC}(\bm x)$ using the current mGP model.
    
    \STATE \textbf{Step 3:} Select the top $B$ points in $\mathcal{X}_{\text{cand}}^{(r)}$ with the highest ALC values. Add them to the training set and remove them from $\mathcal{X}_{\text{cand}}$.
    
    \STATE \textbf{Step 4:} Re-train or update the mGP model using the expanded training dataset.
    
    \STATE \textbf{Step 5:} Evaluate the cross-validation error or training mean squared error (MSE). If the error falls below the threshold \texttt{Tol}, terminate the loop and return the final fitted mGP model.
\ENDFOR
\end{algorithmic}
\end{algorithm}

\section{Examples} \label{sec:examples}

We evaluate the proposed \emph{Active Learning Manifold Gaussian Process (ALmGP)} framework on four benchmark experiments. These experiments are designed to assess the effectiveness of ALmGP under varying geometric structures and data complexities. The acquisition strategy is based on the ALC criterion and model optimization is carried out using the L-BFGS algorithm implemented in the PyTorch library \shortcite{paszke_pytorch_2019}. The optimizer employs strong Wolfe line search conditions (\texttt{line\_search\_fn = strong\_wolfe}), with early stopping triggered when the relative RMSE change
\[
\text{Relative RMSE Change} = \left|\frac{\mathcal{L}_{\text{current}} - \mathcal{L}_{\text{previous}}}{\mathcal{L}_{\text{current}} - \mathcal{L}_{\text{initial}}}\right|
\]
falls below a predefined tolerance threshold. To ensure positivity and improve stability during optimization, all hyperparameters of the mGP model are squared before being passed to the optimizer. All neural network weights are initialized using PyTorch’s default scheme, while GP hyperparameters—output variance and length scale—are initialized to 1.

Model performance is evaluated using the Root Mean Square Error (RMSE) computed on an independently generated test set, which has $n_{\text{test}}$ number of points sampled from $\mathcal{X}$ using Latin Hypercube Design:
\[
\text{RMSE} = \sqrt{\frac{1}{n_{\text{test}}} \sum_i (\hat{y}_i - y_i)^2},
\]
where $\hat{y}_i$ and $y_i$ denote the predicted and ground truth values at test location $\bm x_i$, respectively.
Each experiment is repeated 10 times to ensure statistical robustness. 
The mean and range of RMSE values are reported and compared against a baseline strategy using completely random sampling $B$ design points from $\mathcal{X}_{\text{cand}}$. 

\subsection{Piecewise Trigonometric Function}

\begin{figure}[!h]
    \centering
    \begin{subfigure}{0.48 \textwidth}
        \centering
        \includegraphics[width=\textwidth]{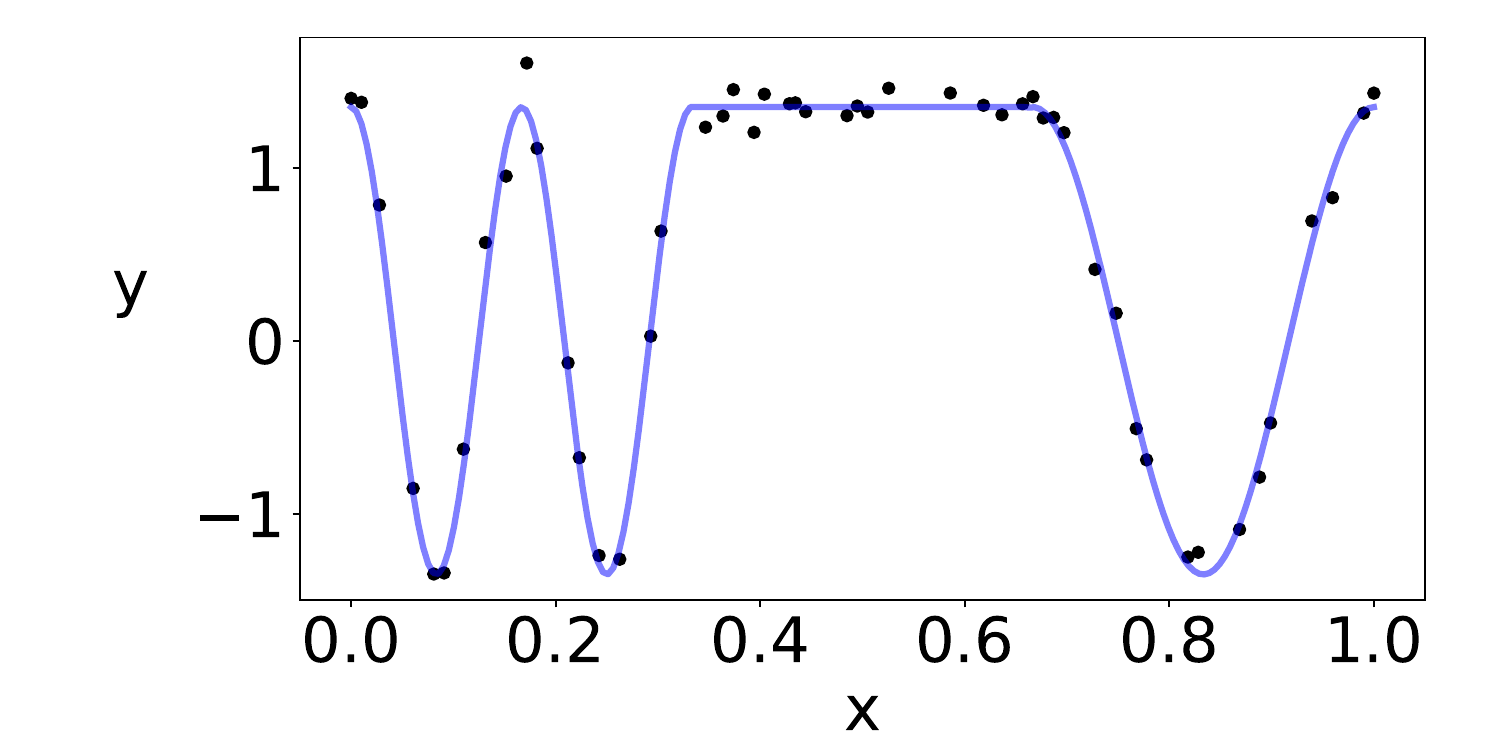}
        \caption{}
    \end{subfigure}
    \hfill
    \begin{subfigure}{0.48 \textwidth}
        \centering
        \includegraphics[width=\textwidth]{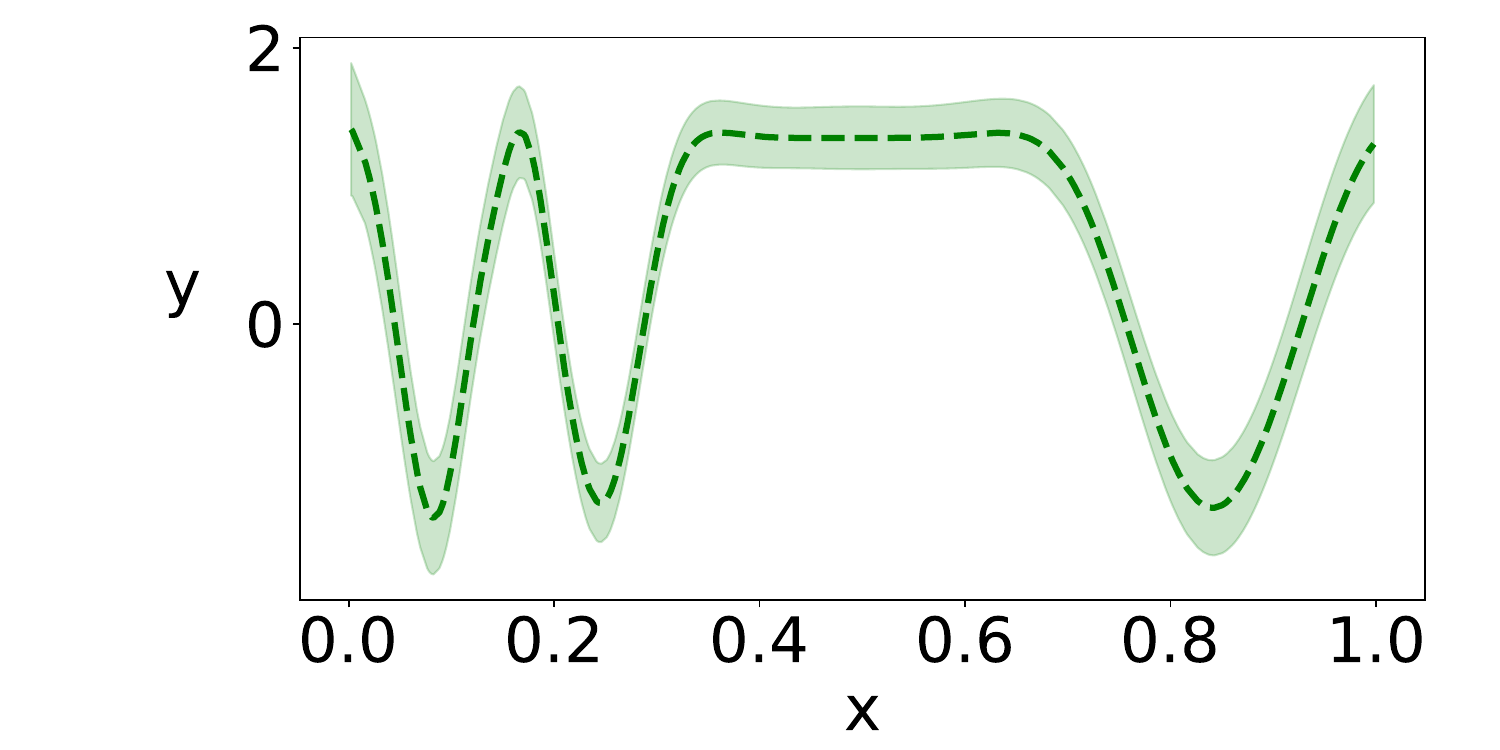}
        \caption{}
    \end{subfigure}\\
    \begin{subfigure}{0.48 \textwidth}
        \centering
        \includegraphics[width=\textwidth]{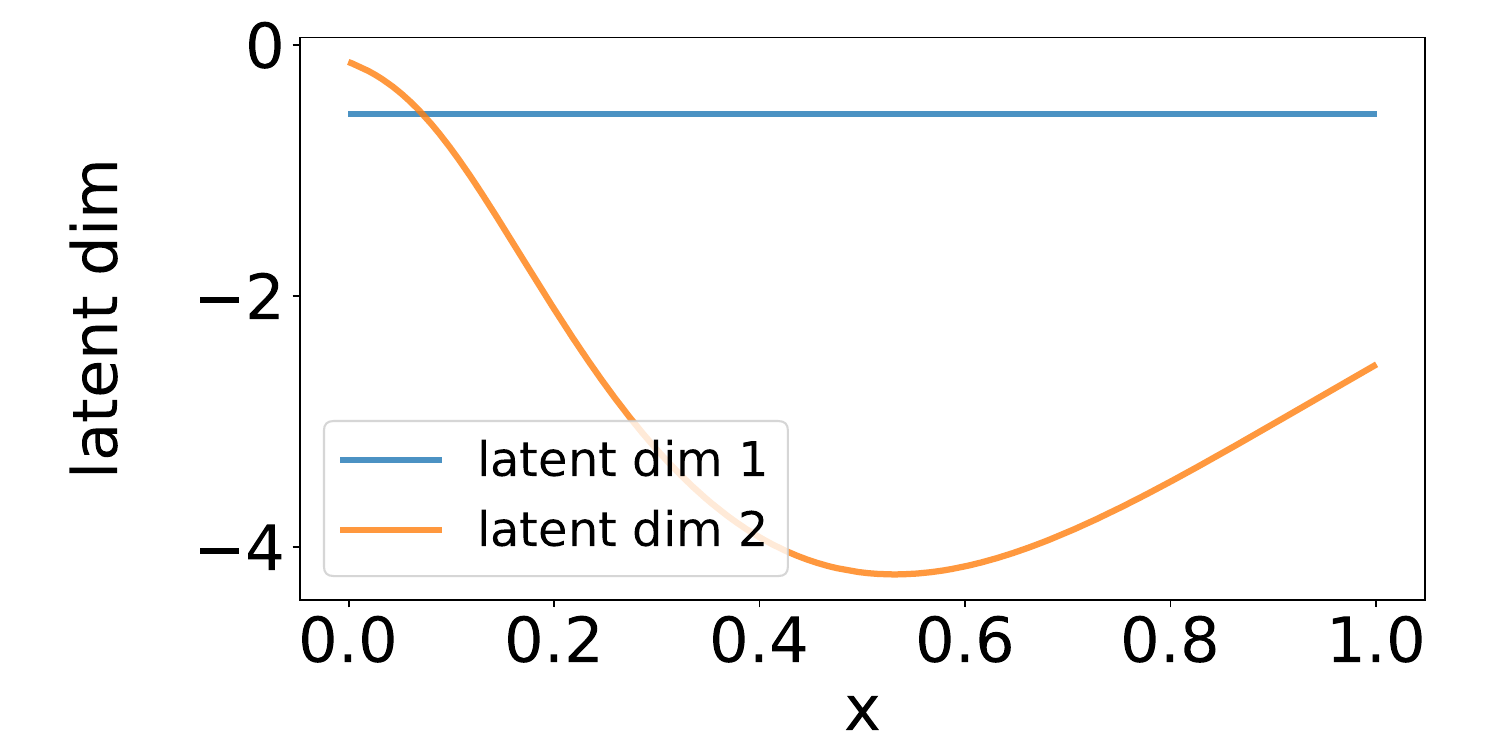}
        \caption{}
    \end{subfigure}
    \hfill
    \begin{subfigure}{0.48 \textwidth}
        \centering
        \includegraphics[width=\textwidth]{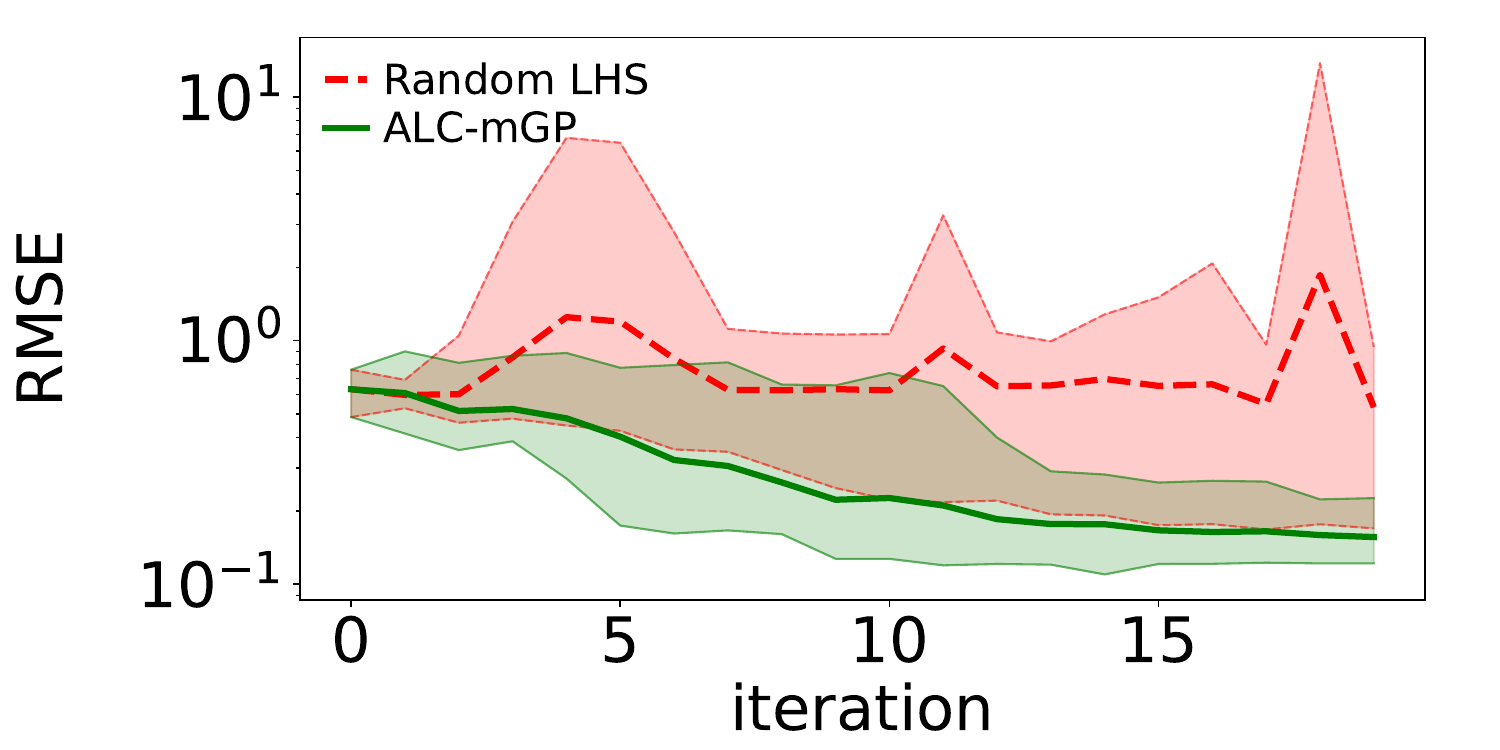}
        \caption{}
    \end{subfigure}
    \caption{(a) The black dots indicate the final selected training data and the blue solid line is the true piecewise trigonometric function. (b) The green dashed line represents the final predicted value the green shaded area denotes the prediction confidence interval.  (c) Learned a two-dimensional latent space. (d) Comparison of test RMSE over iterations: our method (\textcolor{green}{green}, solid) vs.\ random acquisition (\textcolor{red}{red}, dashed). Lines show the mean of 10 runs; shaded areas indicate the minimum to maximum range. \label{fig:trig}}
\end{figure}

This example features a piecewise trigonometric function adapted from \shortciteN{Sauer02012023}, with noise $\epsilon \sim \mathcal{N}(0, 0.1^2)$:
\[
F(x) =
\begin{cases}
1.35\cos(12\pi x), & x \in [0, 0.33], \\
1.35, & x \in [0.33, 0.66], \\
1.35\cos(6\pi x), & x \in [0.66, 1].
\end{cases}
\]
The initial dataset consists of $n_0 = 10$ points generated using Latin Hypercube Design over $[0,1]$. The test set contains 500 uniformly spaced points, and both $\mathcal{X}_{\text{cand}}$ and $\mathcal X_{\text{ref}}$ consist of 100 evenly spaced points in the same interval. The neural network uses $[1\text{-}6\text{-}2]$ architecture and LogSigmoid activation.
The batch size is $B = 1$, and the total budget is $N_{\max} = 50$. L-BFGS uses \texttt{history\_size} = 20, a learning rate of $0.001$, a maximal number of iterations per optimization step $20$ and a maximum of 5,000 training iterations with early stopping tolerance \texttt{Tol} = $10^{-5}$.
The active learning process ends with 15 new samples and achieves an average completion RMSE of $0.156$.
Figure \ref{fig:trig} shows the details of the training data, the fitted function, the latent space, and the comparison of RMSE of active learning for mGP using the proposed ALC strategy and random sampling. 

\subsection{Two-Dimensional Deterministic Function}

\begin{figure}[!h]
    \centering
    \begin{subfigure}{0.32 \textwidth}
        \centering
        \includegraphics[width=\textwidth]{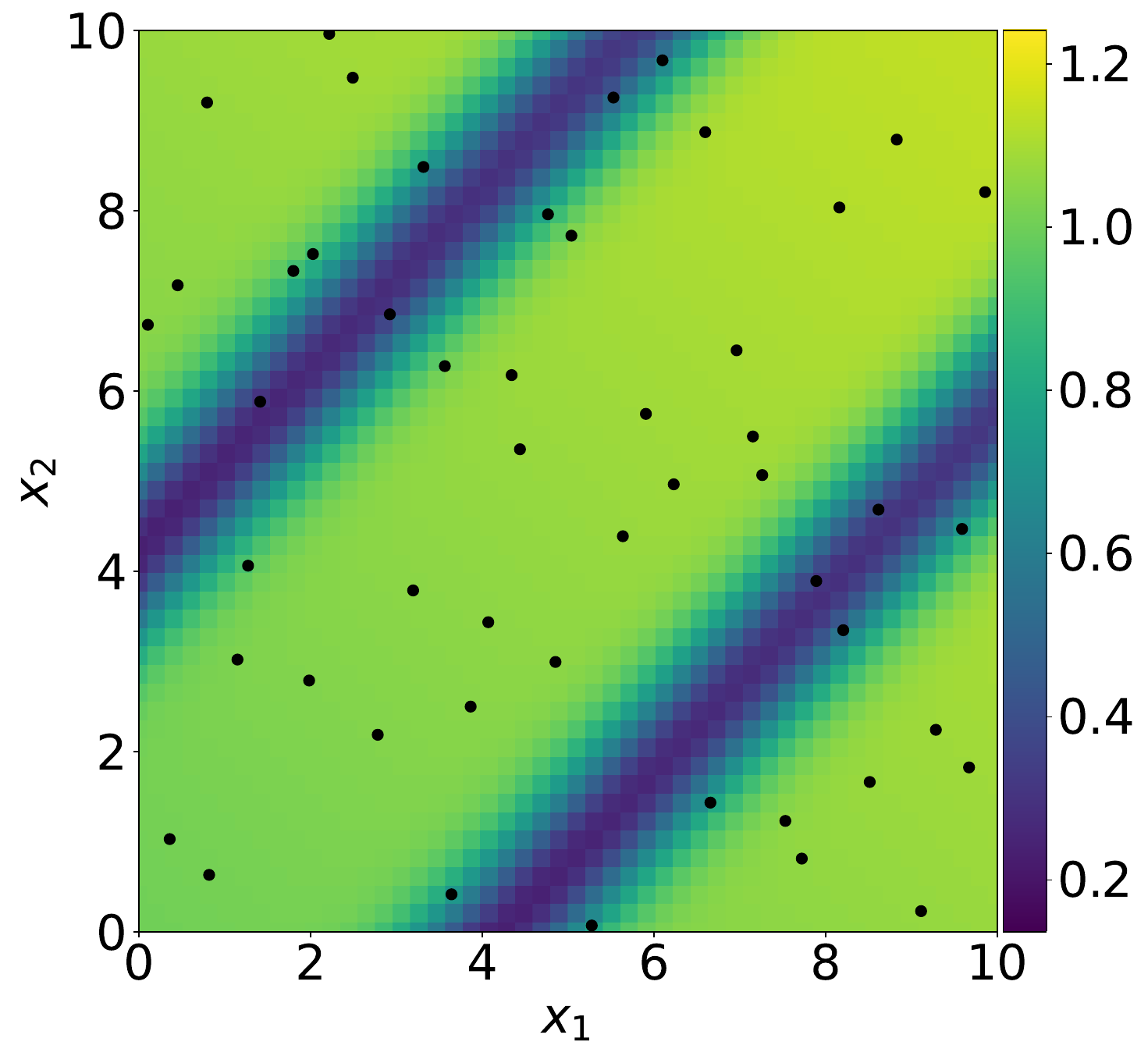}
        \caption{}
    \end{subfigure}
    \begin{subfigure}{0.32 \textwidth}
        \centering
        \includegraphics[width=\textwidth]{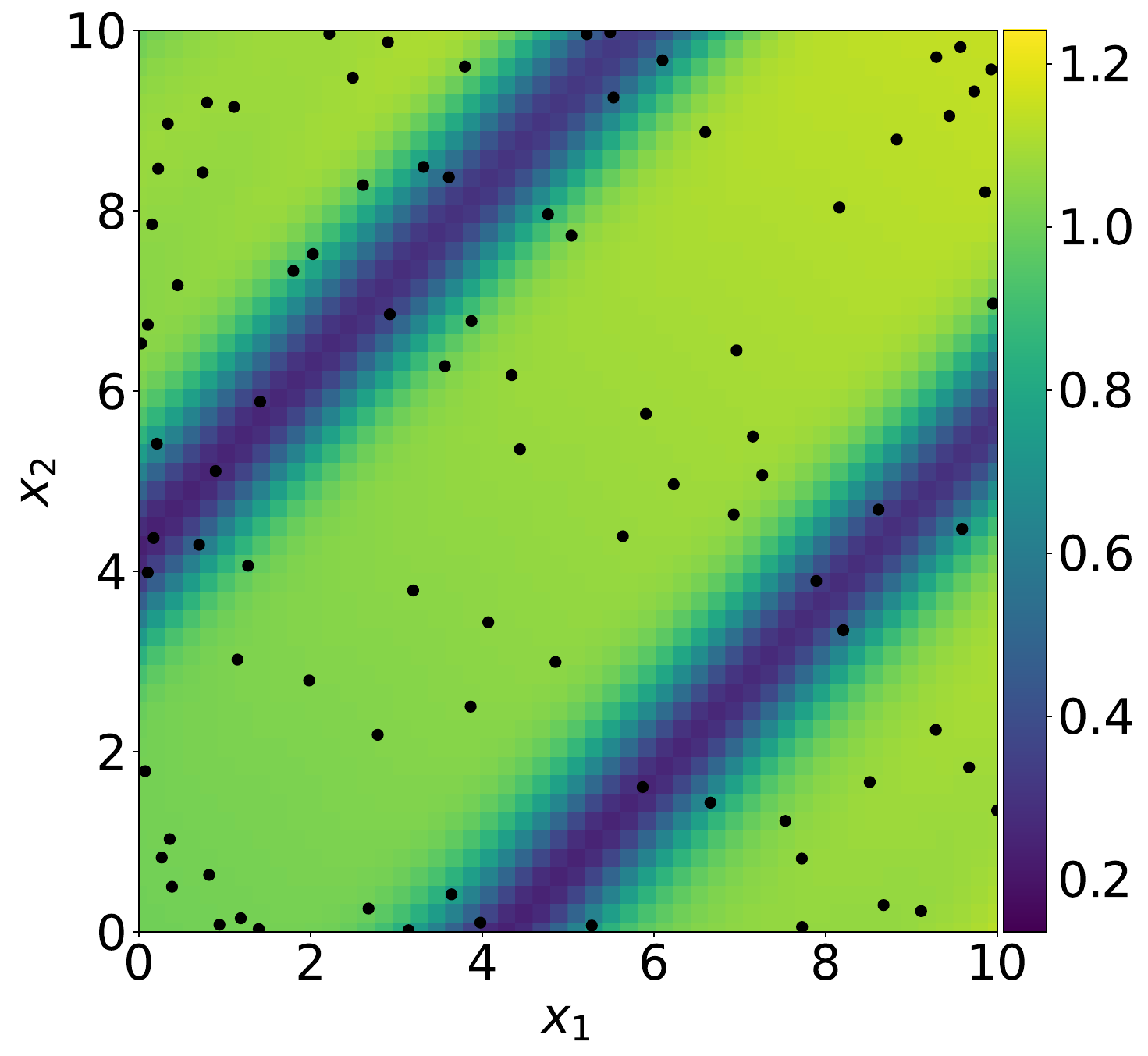}
        \caption{}
    \end{subfigure}
    \begin{subfigure}{0.32 \textwidth}
        \centering
        \includegraphics[width=\textwidth]{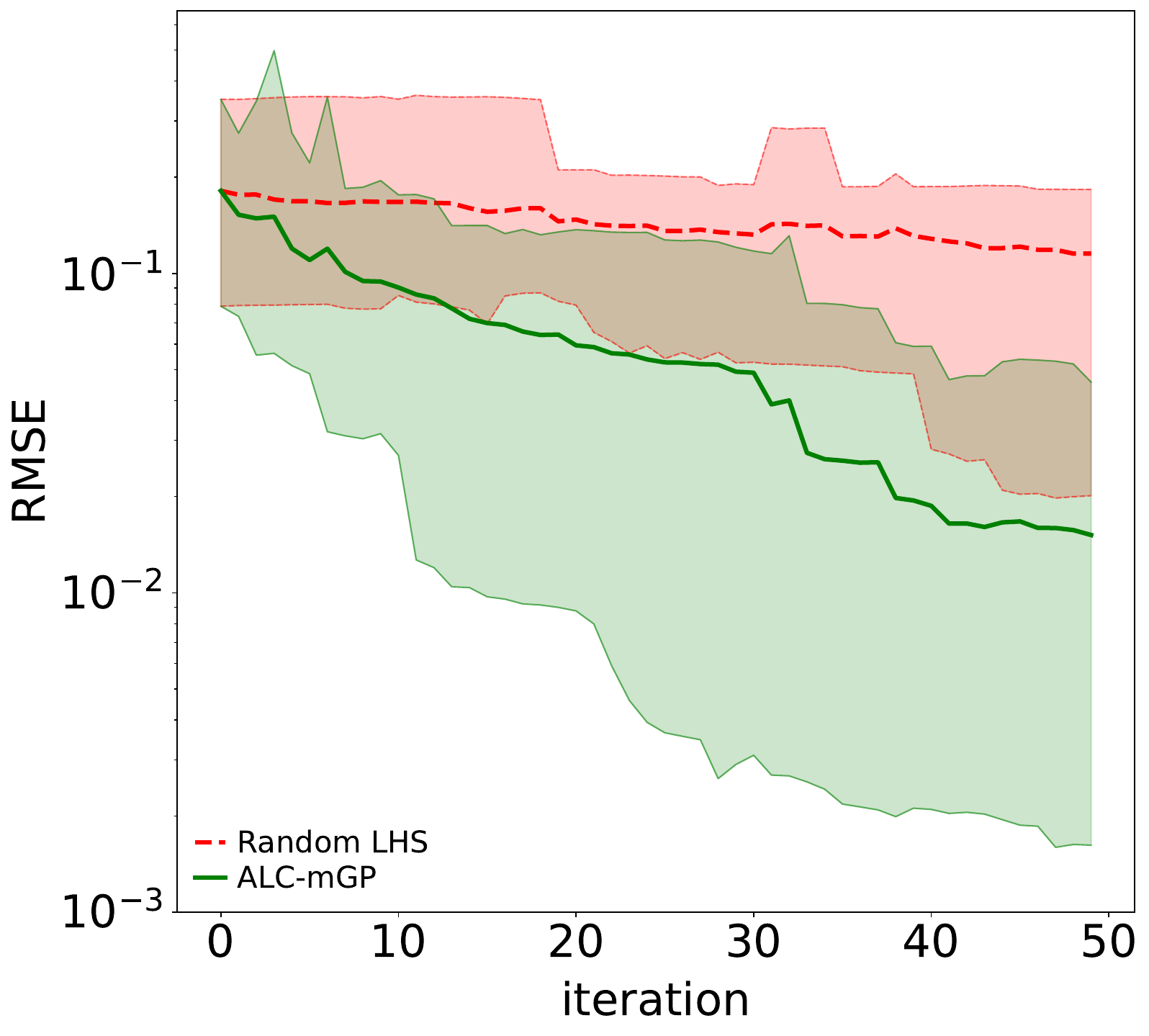}
        \caption{}
    \end{subfigure}
    \caption{(a) The heat map is the true function value with the black dots representing the initial training data. (b) The heat map is the final predicted value, and the black dots are the overall training data. (c) Comparison of test RMSE over iterations: our method (\textcolor{green}{green}, solid) vs.\ random acquisition (\textcolor{red}{red}, dashed). Lines show the mean of 10 runs; shaded areas indicate the minimum to maximum range.\label{fig:two-dim1}}
\end{figure}

\begin{figure}[!h]
    \centering
    \begin{subfigure}{0.31 \textwidth}
        \centering
        \includegraphics[width=\textwidth]{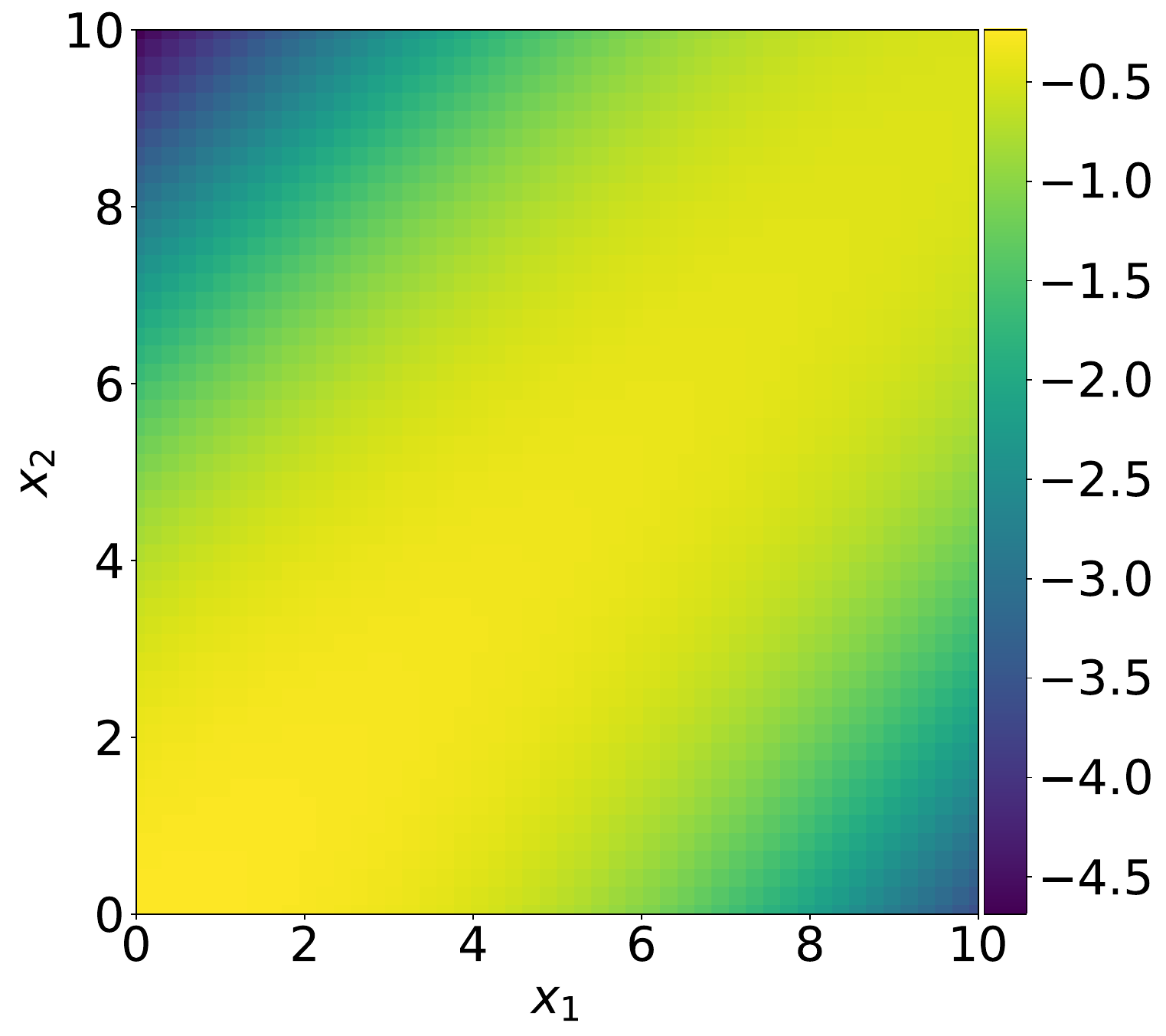}
    \end{subfigure}
    \begin{subfigure}{0.31 \textwidth}
        \centering
        \includegraphics[width=\textwidth]{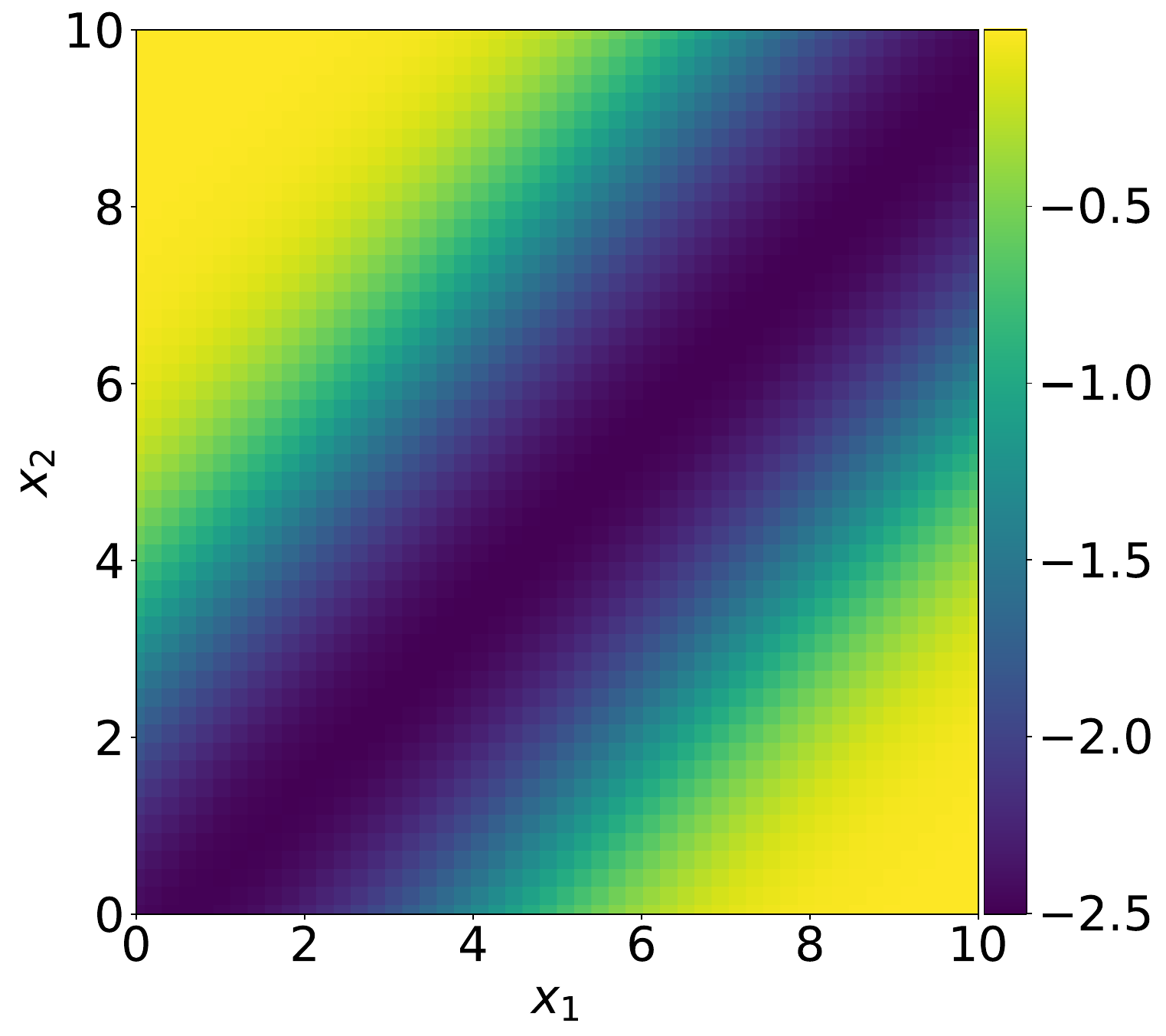}
    \end{subfigure}
    \begin{subfigure}{0.31 \textwidth}
        \centering
        \includegraphics[width=\textwidth]{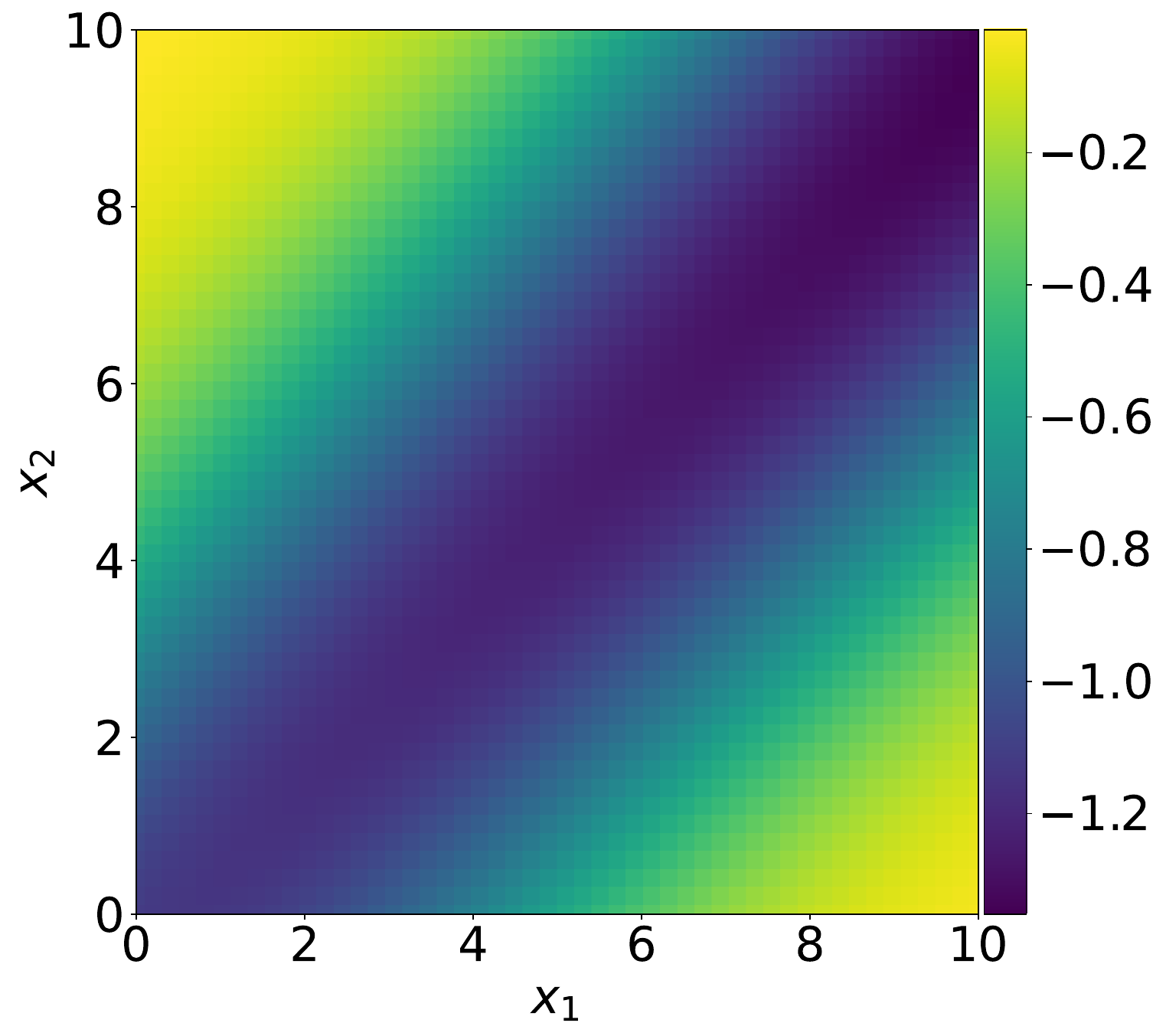}
    \end{subfigure}
    \caption{The three heat maps represent three latent dimensions with respect to the original input $(x_1, x_2)$.\label{fig:two-dim2}}
\end{figure}

We consider a two-dimensional function also from \shortcite{calandra_manifold_2016}, defined as:
\[
f(x_1, x_2) = 1 - \phi(x_2; 3, 0.5^2) - \phi(x_2; -3, 0.5^2) + \frac{x_1}{100},
\]
which is then rotated by $45^{\circ}$. The function $\phi(\cdot;\mu,\sigma^2)$ is the PDF of $\mathcal{N}(\mu,\sigma^2)$.
Then initial training data ($n_0 = 50$), test data, and the reference set $\mathcal X_{\text{ref}}$ are all of size 500 and sampled from $[0, 10]^2$ using LHD. Predictions are also evaluated on a grid over $\mathcal{X}$ with mesh size 0.2.
This neural network maps the 2D input to a latent space, and the architecture is $[2-10-3]$, with the same LogSigmoid activation. The active learning proceeds with batch size $B = 1$ and budget $N_{\max} = 50$. 
The L-BFGS is run with \texttt{history\_size} = 50, learning rate $0.01$, a maximal number of iterations per optimization step $50$, and a max of 5000 iterations.
The active learning procedure concludes with 100 samples after 50 iterations, reaching an average RMSE of $0.0152$ for the proposed method.
Figure \ref{fig:two-dim1} and \ref{fig:two-dim2} show the detailed results. 

\subsection{Three-Dimensional Function on the Unit Sphere}

\begin{figure}[!h]
    \centering
    \begin{subfigure}{0.32 \textwidth}
        \centering
        \includegraphics[width=\textwidth]{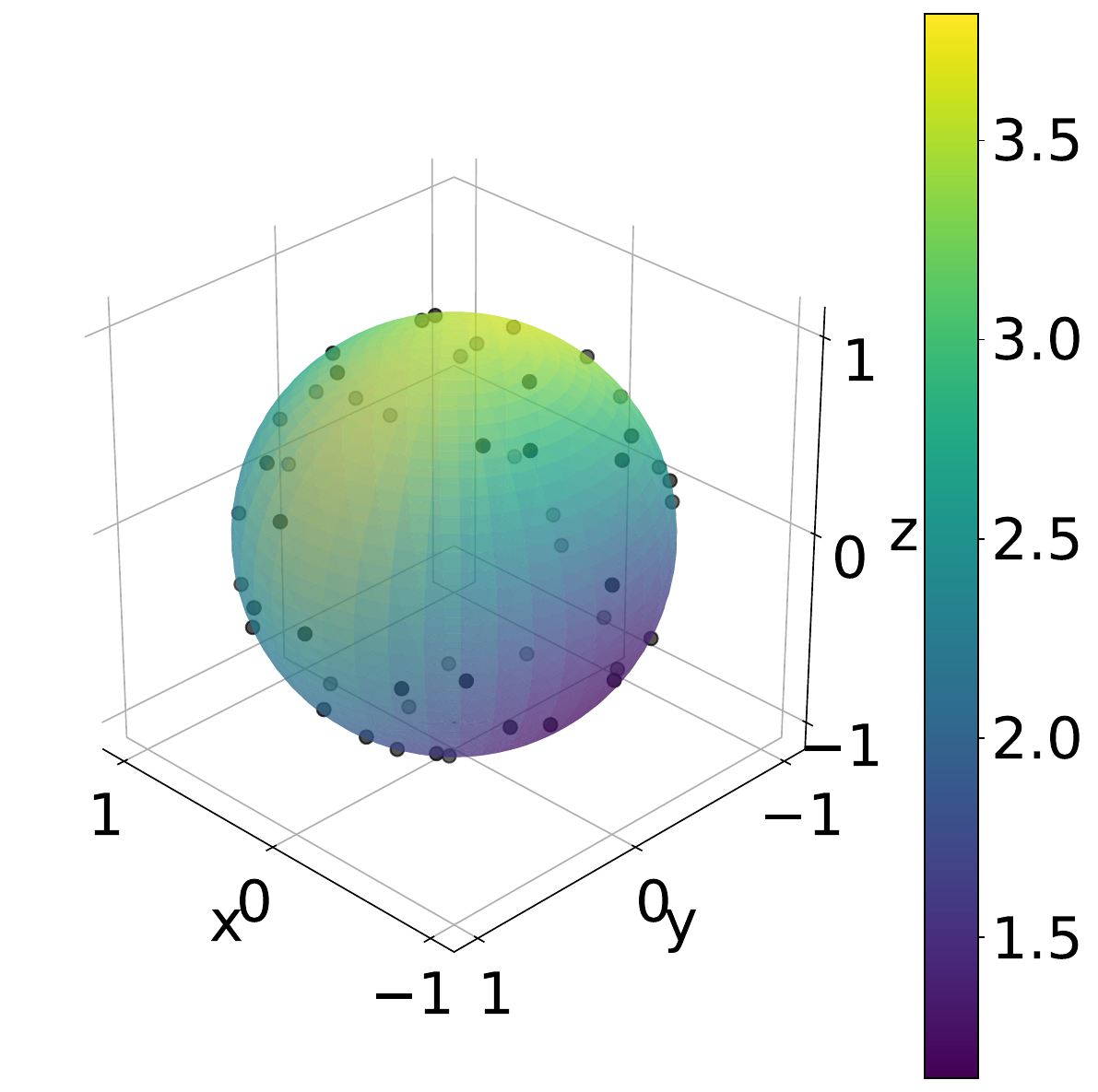}
        \caption{}
    \end{subfigure}
    \hfill
    \begin{subfigure}{0.32 \textwidth}
        \centering
        \includegraphics[width=\textwidth]{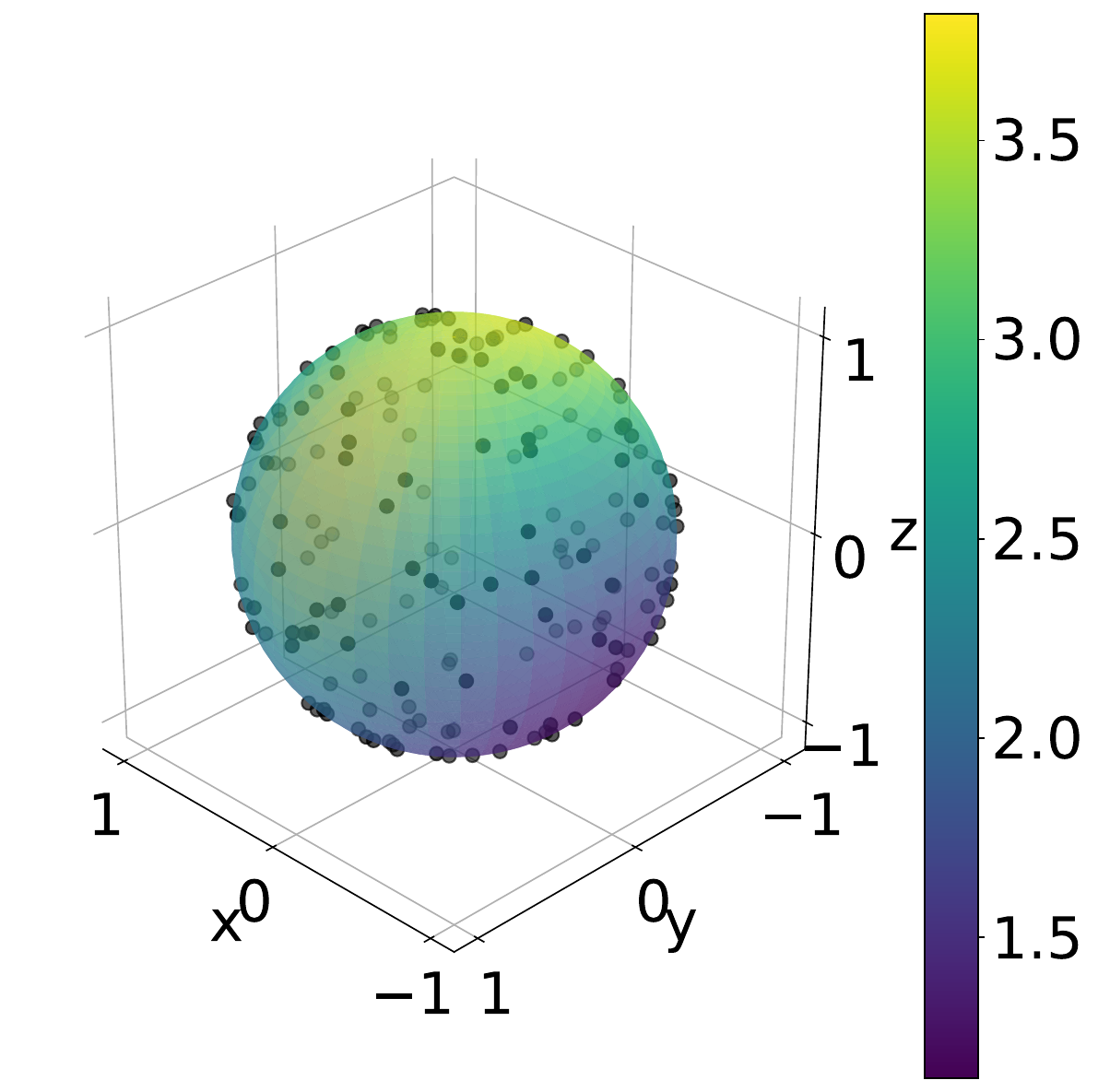}
        \caption{}
    \end{subfigure}
    \hfill
    \begin{subfigure}{0.32 \textwidth}
        \centering
        \includegraphics[width=\textwidth]{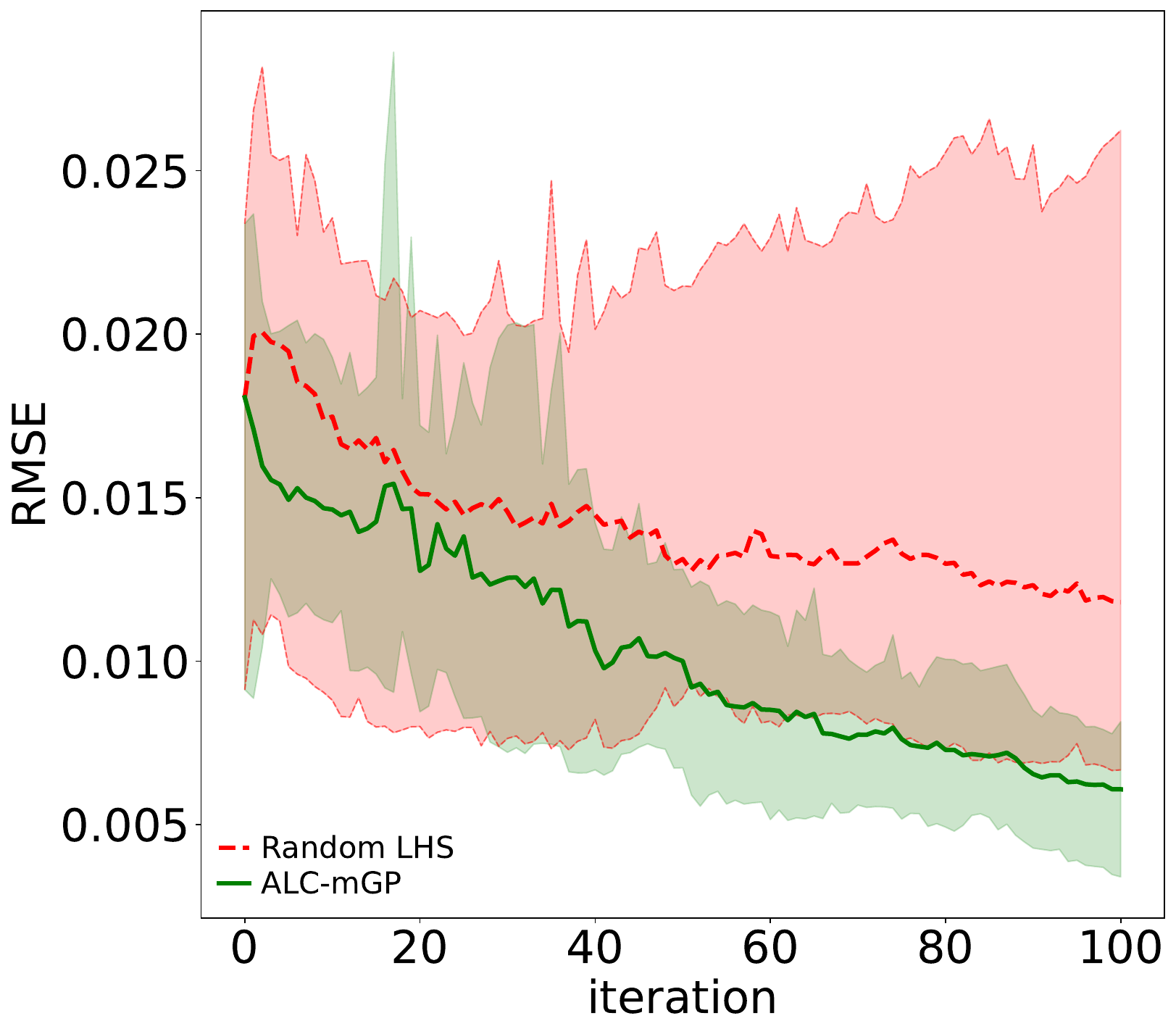}
        \caption{}
    \end{subfigure}
    \caption{(a) The heat map represents the true function value on the 3-dim unit sphere and the dots are the initial training data. (b) The heat map is the final predicted value and the black dots are the overall training data. (c) Comparison of test RMSE over iterations: our method (\textcolor{green}{green}, solid) vs.\ random acquisition (\textcolor{red}{red}, dashed). Lines show the mean of 10 runs; shaded areas indicate the minimum to maximum range. \label{fig:3d1}}
\end{figure}

\begin{figure}[!h]
    \centering
    \begin{subfigure}{0.32 \textwidth}
        \centering
        \includegraphics[width=\textwidth]{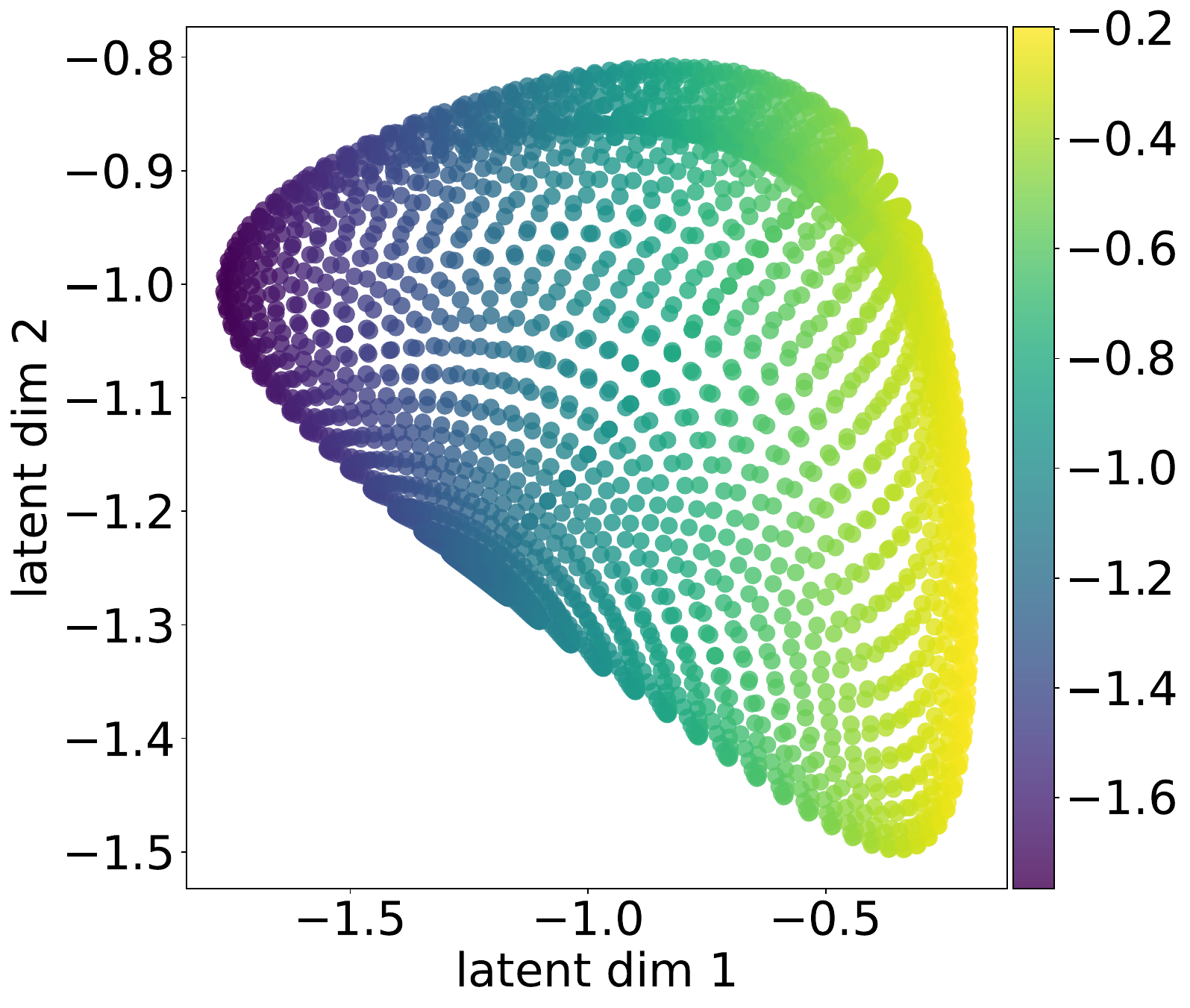}
        \caption{}
    \end{subfigure}
    \begin{subfigure}{0.32 \textwidth}
        \centering
        \includegraphics[width=\textwidth]{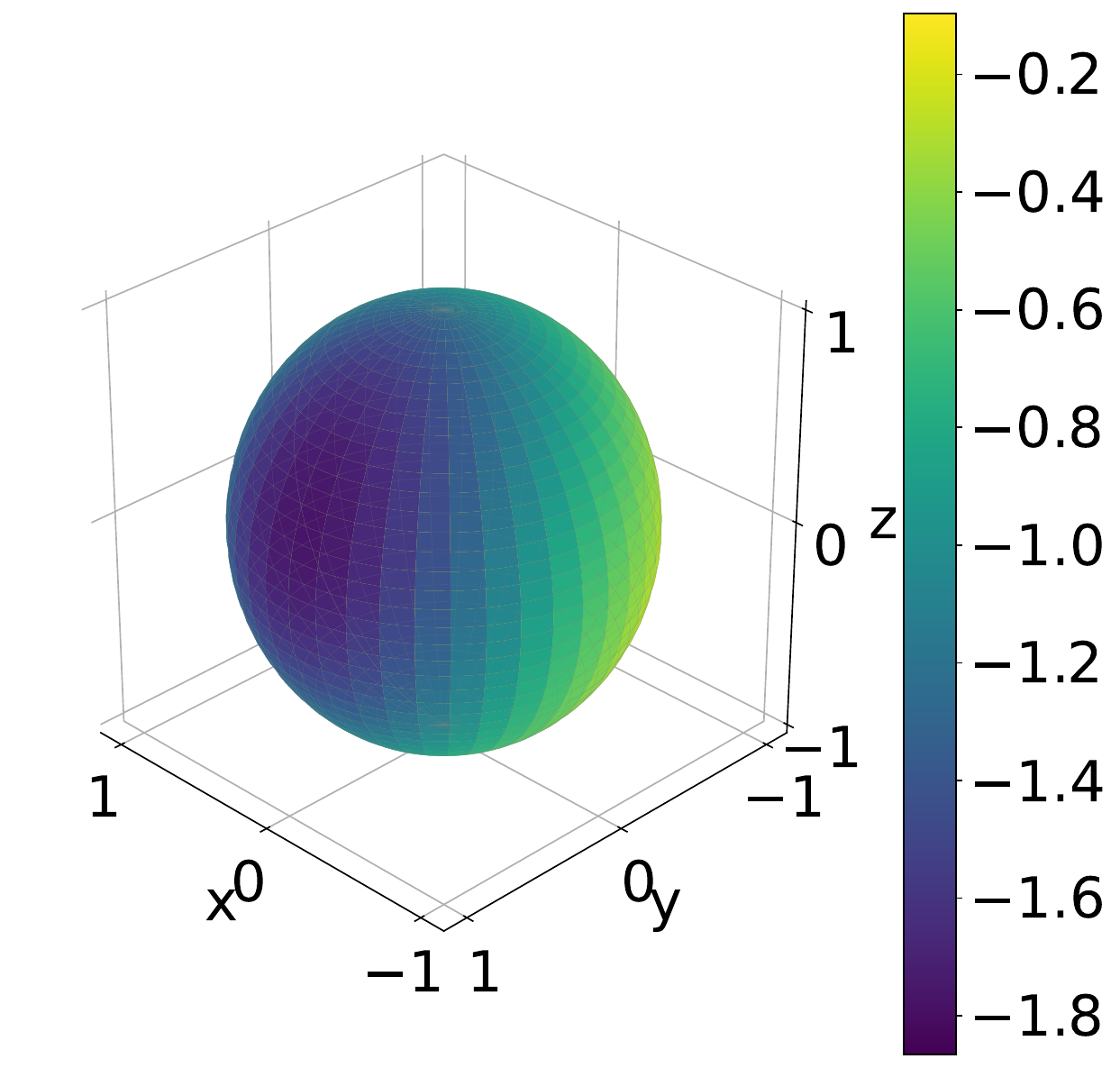}
        \caption{}
    \end{subfigure}
    \begin{subfigure}{0.32 \textwidth}
        \centering
        \includegraphics[width=\textwidth]{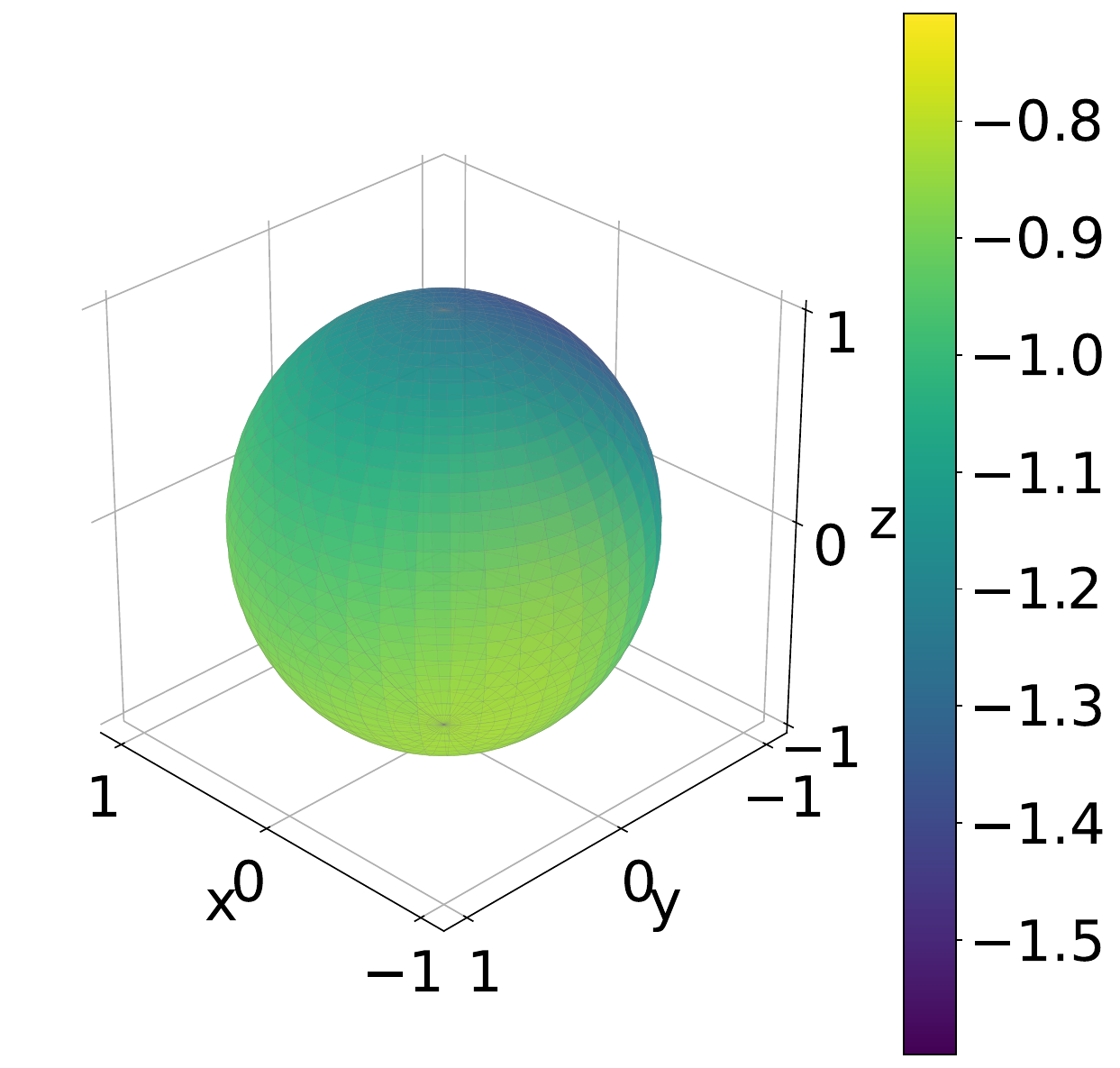}
\caption{}
\end{subfigure}
    \caption{(a) The heat map represents the true function value $f$  with respect to the two latent dimensions. (b) The heat map represents the 1st latent dimension with respect to $(x,y,z)$. (c) The heat map represents the 2nd latent dimension with respect to $(x,y,z)$.\label{fig:3d2}}
\end{figure}

This example uses a test function defined on the 3D unit sphere:
\[
f(x,y,z) = \cos(x) + y^2 + e^z, \quad \text{where } x^2 + y^2 + z^2 = 1.
\]
It can be reformulated on the disk as $g(x,y) = \cos(x) + y^2 + e^{1-x^2-y^2}$ for $x^2 + y^2 \leq 1$.
We construct training, test, and reference sets (each of size 500) by sampling $(v,\alpha) \in [-1, 1] \times [0, 2\pi]$ using Latin Hypercube Design and mapping them to $(x, y, z)$ via:
\[
z = v,\quad x = \sqrt{1 - v^2}\cos(\alpha),\quad y = \sqrt{1 - v^2}\sin(\alpha).
\]
The neural network has architecture $[3\text{-}10\text{-}2]$, and training starts from $n_0 = 50$ samples. 
Active learning is run with $N_{\max} = 100$ and $B = 1$. 
The L-BFGS is configured with \texttt{history\_size} = 50, a learning rate of $0.01$, a maximal number of iterations per optimization step $20$ and 5,000 training iterations. 
The active learning stops when it reaches the full budget of 100 samples, yielding the averaged RMSE as low as $6 \times 10^{-3}$ for the proposed method.
See the results in Figure \ref{fig:3d1} and \ref{fig:3d2}. 
 
\subsection{Borehole}

This example models the groundwater flow function of the borehole, a well-known benchmark for nonlinear regression. The function depends on eight physical parameters, such as radius and aquifer transitivity. The detailed definition of the variable and the math expression of the function are included in \shortciteN{kang2023energetic}. Inputs are generated via LHS and mapped to physical ranges.

\begin{figure}[!h]
    \centering
    \includegraphics[width=0.4\textwidth]{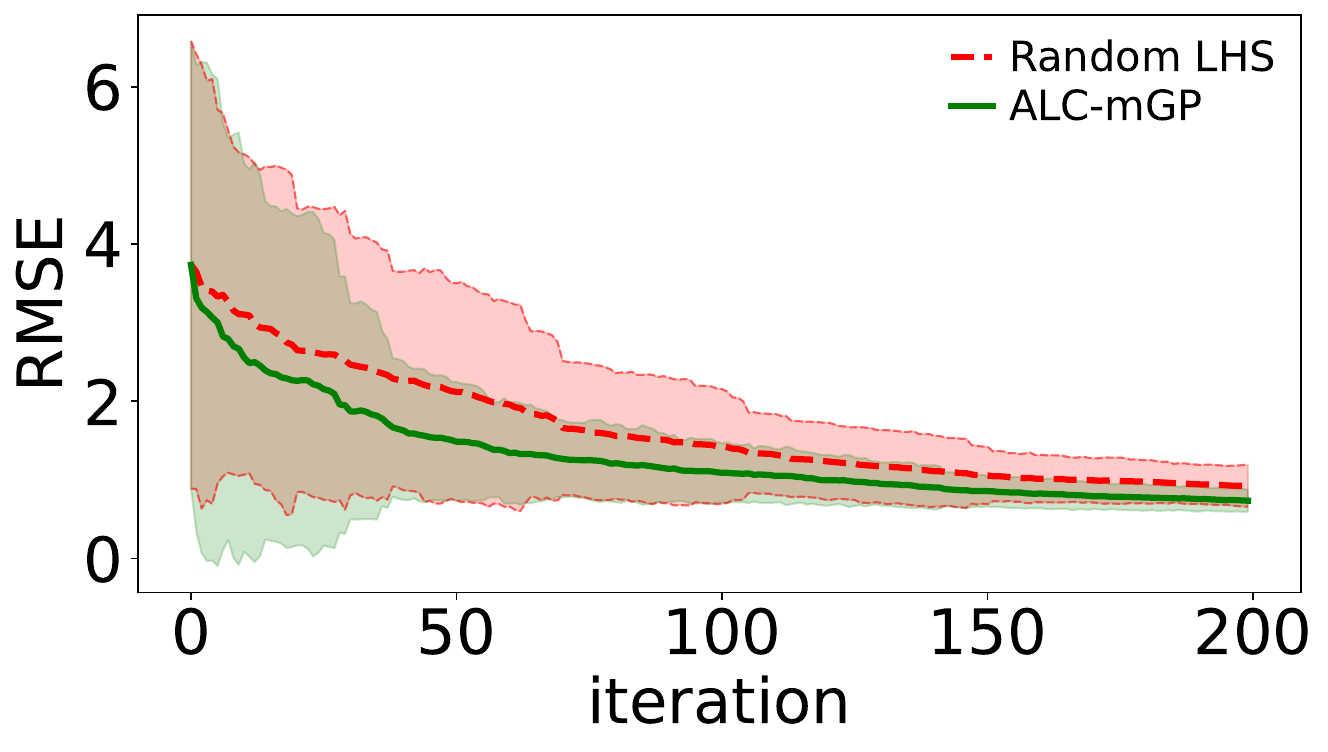}
    \caption{ Comparison of test RMSE over iterations: our method (\textcolor{green}{green}, solid) vs.\ random acquisition (\textcolor{red}{red}, dashed). Lines show the mean of 10 runs; shaded areas indicate the minimum to maximum range.\label{fig:borehole}}
\end{figure}

The neural network has architecture $[8\text{-}30\text{-}4]$, and training starts from $n_0 = 50$ samples. 
Active learning is run with $N_{\max} = 150$ and $B = 1$. 
The L-BFGS uses \texttt{history\_size} = 50, learning rate $0.001$, a maximal number of iterations per optimization step $100$, and 10,000 training iterations. 
Early stopping is triggered when the relative test RMSE change is below $10^{-8}$. 
The proposed active learning stops when it reaches the full budget of 150 samples, yielding an average RMSE as low as $0.6$. See the results in Figure \ref{fig:borehole}. 

\section{Conclusion} \label{sec:conclusion}

This paper introduced an active learning framework for manifold Gaussian Process (mGP) regression, combining the strengths of manifold learning with sequential experimental design. 
By integrating the Active Learning Cohn criterion with mGPs, we demonstrated how to strategically select training points that simultaneously improve predictive accuracy and exploit low-dimensional data structure. 
The proposed method addresses key limitations of standard GPs in high-dimensional spaces, where traditional covariance functions often fail to capture complex or discontinuous patterns. 
Our experiments validated the framework’s efficacy in reducing integrated mean squared prediction error (IMSE) while maintaining computational tractability through efficient hyperparameter optimization and neural network-based manifold learning.

Several promising extensions emerge from this work. First, the current framework assumes a deterministic manifold mapping $M$. 
A Bayesian treatment of $M$ (e.g., via Bayesian neural networks or variational inference) could quantify uncertainty in the latent space, improving robustness when training data is sparse or noisy. This would require advances in approximate inference to handle the non-conjugacy between neural networks and GPs.
Second, the computational cost of active learning scales with the reference set size $m$ is used for IMSE approximation. 
Developing sparse approximations or gradient-based sampling for the ALC integral could enhance scalability to very high-dimensional input spaces. 
Techniques from stochastic optimization, such as mini-batch reference sampling, and merit exploration.
Lastly, the interpretability of learned manifolds could be improved by incorporating domain-specific constraints (e.g., physics-informed neural networks) or disentangled representations. This would bridge the gap between black-box flexibility and interpretable modeling, particularly in scientific applications where mechanistic understanding is paramount.


\section*{ACKNOWLEDGMENTS}
The four authors’ work was partially supported by NSF Grant DMS-2429324. Part of this research was performed while L. Kang was visiting the Institute for Mathematical and Statistical Innovation (IMSI) at the University of Chicago from March 3 to May 24, 2025, which is supported by the National Science Foundation (Grant DMS-1929348).

\bibliographystyle{wsc}
\bibliography{ALGPR,Books,Papers}

\begin{thebibliography}{}

\bibitem[\protect\citeauthoryear{Beyer, Goldstein, Ramakrishnan, and
  Shaft}{Beyer et~al.}{1999}]{beyer_when_1999}
Beyer, K., J.~Goldstein, R.~Ramakrishnan, and U.~Shaft. 1999.
\newblock ``When is “nearest neighbor” meaningful?''.
\newblock In {\em Database Theory—ICDT’99: 7th International Conference
  Jerusalem, Israel, January 10--12, 1999 Proceedings 7},  217--235.
\newblock Springer.

\bibitem[\protect\citeauthoryear{Binois, Huang, Gramacy, and and}{Binois
  et~al.}{2019}]{binois_replication_2019}
Binois, M., J.~Huang, R.~B. Gramacy, and M.~L. and. 2019.
\newblock ``Replication or Exploration? Sequential Design for Stochastic
  Simulation Experiments''.
\newblock {\em
  Technometrics\/}~61(1):7--23~\url{https://doi.org/10.1080/00401706.2018.1469433}.


\bibitem[\protect\citeauthoryear{Borovitskiy, Terenin, Mostowsky, and
  Deisenroth~(he/him)}{Borovitskiy et~al.}{2020}]{borovitskiy_matern_2023}
Borovitskiy, V., A.~Terenin, P.~Mostowsky, and M.~Deisenroth~(he/him). 2020.
\newblock ``Mat\'{e}rn Gaussian Processes on Riemannian Manifolds''.
\newblock In {\em Advances in Neural Information Processing Systems}, edited
  by\ H.~Larochelle, M.~Ranzato, R.~Hadsell, M.~Balcan, and H.~Lin, Volume~33,
  12426--12437.
\newblock IEEE: Curran Associates, Inc.

\bibitem[\protect\citeauthoryear{Calandra, Peters, Rasmussen, and
  Deisenroth}{Calandra et~al.}{2016}]{calandra_manifold_2016}
Calandra, R., J.~Peters, C.~E. Rasmussen, and M.~P. Deisenroth. 2016.
\newblock ``Manifold Gaussian processes for regression''.
\newblock In {\em 2016 International joint conference on neural networks
  (IJCNN)},  3338--3345.
\newblock IEEE~\url{https://doi.org/10.1109/IJCNN.2016.7727626}.

\bibitem[\protect\citeauthoryear{Chen, Kang, and and}{Chen
  et~al.}{2021}]{Chen03072021}
Chen, J., L.~Kang, and G.~L. and. 2021.
\newblock ``Gaussian Process Assisted Active Learning of Physical Laws''.
\newblock {\em
  Technometrics\/}~63(3):329--342~\url{https://doi.org/10.1080/00401706.2020.1817790}.


\bibitem[\protect\citeauthoryear{Cohn}{Cohn}{1993}]{cohn_neural_1996}
Cohn, D. 1993.
\newblock ``Neural Network Exploration Using Optimal Experiment Design''.
\newblock In {\em Advances in Neural Information Processing Systems}, edited
  by\ J.~Cowan, G.~Tesauro, and J.~Alspector, Volume~6,  679--686:
  Morgan-Kaufmann.

\bibitem[\protect\citeauthoryear{Cohn, Ghahramani, and Jordan}{Cohn
  et~al.}{1994}]{cohn_active_1994}
Cohn, D., Z.~Ghahramani, and M.~Jordan. 1994.
\newblock ``Active Learning with Statistical Models''.
\newblock In {\em Advances in Neural Information Processing Systems}, edited
  by\ G.~Tesauro, D.~Touretzky, and T.~Leen, Volume~7.
\newblock IEEE: MIT Press.

\bibitem[\protect\citeauthoryear{Damianou and Lawrence}{Damianou and
  Lawrence}{2013}]{pmlr-v31-damianou13a}
Damianou, A., and N.~D. Lawrence. 2013, 29 Apr--01 May.
\newblock ``Deep {G}aussian Processes''.
\newblock In {\em Proceedings of the Sixteenth International Conference on
  Artificial Intelligence and Statistics}, edited by\ C.~M. Carvalho and
  P.~Ravikumar, Volume~31 of {\em Proceedings of Machine Learning Research},
  207--215.
\newblock Scottsdale, Arizona, USA: PMLR.

\bibitem[\protect\citeauthoryear{Fichera, Borovitskiy, Krause, and
  Billard}{Fichera et~al.}{2023}]{fichera_implicit_2024}
Fichera, B., S.~Borovitskiy, A.~Krause, and A.~G. Billard. 2023.
\newblock ``Implicit Manifold Gaussian Process Regression''.
\newblock In {\em Advances in Neural Information Processing Systems}, edited
  by\ A.~Oh, T.~Naumann, A.~Globerson, K.~Saenko, M.~Hardt, and S.~Levine,
  Volume~36,  67701--67720: Curran Associates, Inc.

\bibitem[\protect\citeauthoryear{Frazier}{Frazier}{2018}]{frazier2018bayesian}
Frazier, P.~I. 2018.
\newblock ``Bayesian optimization''.
\newblock In {\em Recent advances in optimization and modeling of contemporary
  problems},  255--278. Informs~\url{https://doi.org/10.1287/educ.2018.0188}.

\bibitem[\protect\citeauthoryear{Heo and Sung}{Heo and
  Sung}{2025}]{heo2025active}
Heo, J., and C.-L. Sung. 2025.
\newblock ``Active learning for a recursive non-additive emulator for
  multi-fidelity computer experiments''.
\newblock {\em
  Technometrics\/}~67(1):58--72~\url{https://doi.org/10.1080/00401706.2024.2376173}.


\bibitem[\protect\citeauthoryear{Houlsby, Husz{\'a}r, Ghahramani, and
  Lengyel}{Houlsby, Neil and Husz{\'a}r, Ferenc and Ghahramani, Zoubin and
  Lengyel, M{\'a}t{\'e}}{2011}]{houlsby2011bayesian}
Houlsby, Neil and Husz{\'a}r, Ferenc and Ghahramani, Zoubin and Lengyel,
  M{\'a}t{\'e} 2011.
\newblock ``Bayesian active learning for classification and preference
  learning''~\url{https://doi.org/10.48550/arXiv.1112.5745}.

\bibitem[\protect\citeauthoryear{Hung}{Hung}{2011}]{hung2011penalized}
Hung, Y. 2011.
\newblock ``Penalized blind kriging in computer experiments''.
\newblock {\em Statistica
  Sinica\/}~21(3):1171~\url{https://doi.org/10.5705/ss.2009.226}.


\bibitem[\protect\citeauthoryear{Joseph, Hung, and Sudjianto}{Joseph
  et~al.}{2008}]{joseph2008blind}
Joseph, V.~R., Y.~Hung, and A.~Sudjianto. 2008.
\newblock ``Blind kriging: A new method for developing metamodels''.
\newblock {\em Journal of mechanical
  design\/}~130(3):031102~\url{https://doi.org/10.1115/1.2829873}.


\bibitem[\protect\citeauthoryear{Joseph and Kang}{Joseph and
  Kang}{2011}]{joseph2011regression}
Joseph, V.~R., and L.~Kang. 2011.
\newblock ``Regression-based inverse distance weighting with applications to
  computer experiments''.
\newblock {\em
  Technometrics\/}~53(3):254--265~\url{https://doi.org/10.1198/TECH.2011.09154}.


\bibitem[\protect\citeauthoryear{Kang, Cheng, Wang, and Liu}{Kang
  et~al.}{2023}]{kang2023energetic}
Kang, L., Y.~Cheng, Y.~Wang, and C.~Liu. 2023.
\newblock ``Energetic Variational Gaussian Process Regression for Computer
  Experiments''.
\newblock {\em arXiv preprint
  arXiv:2401.00395\/}~\url{https://doi.org/10.48550/arXiv.2401.00395}.


\bibitem[\protect\citeauthoryear{Kang, Cheng, Wang, and Liu}{Kang
  et~al.}{2024}]{kang2024energetic}
Kang, L., Y.~Cheng, Y.~Wang, and C.~Liu. 2024.
\newblock ``Energetic Variational Gaussian Process Regression''.
\newblock In {\em 2024 Winter Simulation Conference (WSC)},  3542--3553.
\newblock INFROMS~\url{https://doi.org/10.1109/WSC63780.2024.10838889}.

\bibitem[\protect\citeauthoryear{Kapoor, Grauman, Urtasun, and Darrell}{Kapoor
  et~al.}{2007}]{kapoor2007active}
Kapoor, A., K.~Grauman, R.~Urtasun, and T.~Darrell. 2007.
\newblock ``Active Learning with Gaussian Processes for Object
  Categorization''.
\newblock In {\em 2007 IEEE 11th International Conference on Computer Vision},
  1--8~\url{https://doi.org/10.1109/ICCV.2007.4408844}.

\bibitem[\protect\citeauthoryear{Kim, Sanz-Alonso, and Yang}{Kim
  et~al.}{2024}]{doi:10.1137/22M1529907}
Kim, H., D.~Sanz-Alonso, and R.~Yang. 2024.
\newblock ``Optimization on Manifolds via Graph Gaussian Processes''.
\newblock {\em SIAM Journal on Mathematics of Data
  Science\/}~6(1):1--25~\url{https://doi.org/10.1137/22M1529907}.


\bibitem[\protect\citeauthoryear{Krause, Singh, and Guestrin}{Krause
  et~al.}{2008}]{krause2008near}
Krause, A., A.~Singh, and C.~Guestrin. 2008.
\newblock ``Near-optimal sensor placements in Gaussian processes: Theory,
  efficient algorithms and empirical studies''.
\newblock {\em Journal of Machine Learning Research\/}~9(2):235--284.


\bibitem[\protect\citeauthoryear{Ma, Li, and Hernandez-Lobato}{Ma
  et~al.}{2019}]{ma2019variational}
Ma, C., Y.~Li, and J.~M. Hernandez-Lobato. 2019, 09--15 Jun.
\newblock ``Variational Implicit Processes''.
\newblock In {\em Proceedings of the 36th International Conference on Machine
  Learning}, edited by\ K.~Chaudhuri and R.~Salakhutdinov, Volume~97 of {\em
  Proceedings of Machine Learning Research},  4222--4233: PMLR.

\bibitem[\protect\citeauthoryear{MacKay}{MacKay}{1992}]{mackay1992information}
MacKay, D. J.~C. 1992, 07.
\newblock ``Information-Based Objective Functions for Active Data Selection''.
\newblock {\em Neural
  Computation\/}~4(4):590--604~\url{https://doi.org/10.1162/neco.1992.4.4.590}.


\bibitem[\protect\citeauthoryear{Mallasto and Feragen}{Mallasto and
  Feragen}{2018}]{mallasto_wrapped_2018}
Mallasto, A., and A.~Feragen. 2018, June.
\newblock ``Wrapped Gaussian Process Regression on Riemannian Manifolds''.
\newblock In {\em Proceedings of the IEEE Conference on Computer Vision and
  Pattern Recognition (CVPR)}.

\bibitem[\protect\citeauthoryear{Martinez-Cantin, Tee, and
  McCourt}{Martinez-Cantin et~al.}{2018}]{pmlr-v84-martinez-cantin18a}
Martinez-Cantin, R., K.~Tee, and M.~McCourt. 2018, 09--11 Apr.
\newblock ``Practical Bayesian optimization in the presence of outliers''.
\newblock In {\em Proceedings of the Twenty-First International Conference on
  Artificial Intelligence and Statistics}, edited by\ A.~Storkey and
  F.~Perez-Cruz, Volume~84 of {\em Proceedings of Machine Learning Research},
  1722--1731: PMLR.

\bibitem[\protect\citeauthoryear{Meilă and Zhang}{Meilă and
  Zhang}{}]{meila_manifold_2024}
Meilă, M., and H.~Zhang.
\newblock ``Manifold {{Learning}}: {{What}}, {{How}}, and {{Why}}''.
\newblock
  ~11:393--417~\url{https://doi.org/10.1146/annurev-statistics-040522-115238}.


\bibitem[\protect\citeauthoryear{Nocedal and Wright}{Nocedal and
  Wright}{2006}]{nocedal_numerical_2006}
Nocedal, J., and S.~J. Wright. 2006.
\newblock {\em Numerical Optimization\/}. 2nd ed ed.
\newblock Springer Series in Operations Research. Springer.
 \begin{quotation}\noindent OCLC: ocm68629100 \end{quotation}

\bibitem[\protect\citeauthoryear{Paszke, Gross, Massa, Lerer, Bradbury, Chanan,
  Killeen, Lin, Gimelshein, Antiga, Desmaison, Kopf, Yang, DeVito, Raison,
  Tejani, Chilamkurthy, Steiner, Fang, Bai, and Chintala}{Paszke
  et~al.}{2019}]{paszke_pytorch_2019}
Paszke, A., S.~Gross, F.~Massa, A.~Lerer, J.~Bradbury, G.~Chanan,  {\em et~al}.
  2019.
\newblock ``PyTorch: An Imperative Style, High-Performance Deep Learning
  Library''.
\newblock In {\em Advances in Neural Information Processing Systems}, edited
  by\ H.~Wallach, H.~Larochelle, A.~Beygelzimer, F.~d\textquotesingle
  Alch\'{e}-Buc, E.~Fox, and R.~Garnett, Volume~32: Curran Associates, Inc.

\bibitem[\protect\citeauthoryear{Rasmussen and Williams}{Rasmussen and
  Williams}{2006}]{rasmussen2006gaussian}
Rasmussen, C.~E., and C.~K.~I. Williams. 2006.
\newblock {\em Gaussian Processes for Machine Learning}.
\newblock MIT Press.


\bibitem[\protect\citeauthoryear{Santner, Williams, Notz, and Williams}{Santner
  et~al.}{2003}]{santner2003design}
Santner, T.~J., B.~J. Williams, W.~I. Notz, and B.~J. Williams. 2003.
\newblock {\em The design and analysis of computer experiments}, Volume~1.
\newblock New York: Springer.


\bibitem[\protect\citeauthoryear{Sauer, Gramacy, and and}{Sauer
  et~al.}{2023}]{Sauer02012023}
Sauer, A., R.~B. Gramacy, and D.~H. and. 2023.
\newblock ``Active Learning for Deep Gaussian Process Surrogates''.
\newblock {\em
  Technometrics\/}~65(1):4--18~\url{https://doi.org/10.1080/00401706.2021.2008505}.


\bibitem[\protect\citeauthoryear{Seo, Wallat, Graepel, and Obermayer}{Seo
  et~al.}{}]{seo_gaussian_2000}
Seo, S., M.~Wallat, T.~Graepel, and K.~Obermayer.
\newblock ``Gaussian Process Regression: Active Data Selection and Test Point
  Rejection''.
\newblock In {\em Proceedings of the {{IEEE-INNS-ENNS International Joint
  Conference}} on {{Neural Networks}}. {{IJCNN}} 2000. {{Neural Computing}}:
  {{New Challenges}} and {{Perspectives}} for the {{New Millennium}}},
  Volume~3,  241--246 vol.3~\url{https://doi.org/10.1109/IJCNN.2000.861310}.

\bibitem[\protect\citeauthoryear{Seo, Wallat, Graepel, and Obermayer}{Seo
  et~al.}{2000}]{seo2000gaussian}
Seo, S., M.~Wallat, T.~Graepel, and K.~Obermayer. 2000.
\newblock ``Gaussian process regression: active data selection and test point
  rejection''.
\newblock In {\em Proceedings of the IEEE-INNS-ENNS International Joint
  Conference on Neural Networks. IJCNN 2000. Neural Computing: New Challenges
  and Perspectives for the New Millennium}, Volume~3,  241--246
  vol.3~\url{https://doi.org/10.1109/IJCNN.2000.861310}.

\bibitem[\protect\citeauthoryear{Srinivas, Krause, Kakade, and Seeger}{Srinivas
  et~al.}{2010}]{srinivas2010gaussian}
Srinivas, N., A.~Krause, S.~Kakade, and M.~Seeger. 2010.
\newblock ``Gaussian Process Optimization in the Bandit Setting: No Regret and
  Experimental Design''.
\newblock In {\em ICML'10: Proceedings of the 27 th International Conference on
  Machine Learning},  1015–1022.
\newblock MadisonWIUnited States: Omnipress.

\bibitem[\protect\citeauthoryear{Tenenbaum, Silva, and Langford}{Tenenbaum
  et~al.}{2000}]{tenenbaum_global_2000}
Tenenbaum, J.~B., V.~d. Silva, and J.~C. Langford. 2000.
\newblock ``A global geometric framework for nonlinear dimensionality
  reduction''.
\newblock {\em
  science\/}~290(5500):2319--2323~\url{https://doi.org/10.1126/science.290.5500.2319}.


\end{thebibliography}

\section*{AUTHOR BIOGRAPHIES}
\noindent {\bf \MakeUppercase{Yuanxing Cheng}} is a Ph.D student in the Department of Applied Mathematics at the Illinois Institute of Technology in Chicago, IL. Advised by Dr. Lulu Kang and Dr. Chun Liu, he has worked on the topic of uncertainty quantification. His email address is \email{ycheng46@hawk.iit.edu}.\\

\noindent {\bf \MakeUppercase{Lulu Kang}} is an Associate Professor in the Department of Mathematics and Statistics at the University of Massachusetts Amherst. Her research interests include Statistical Learning/Machine Learning, Statistical Design of Experiments, Uncertainty Quantification, Bayesian Statistical Modeling, Approximate Inference, and Optimization. She serves as an associate editor for \emph{Technometrics} and \emph{SIAM/ASA Journal on Uncertainty Quantification}. Her email address is  \email{lulukang@umass.edu} and her website is \url{https://sites.google.com/umass.edu/lulukang/home}. \\

\noindent {\bf \MakeUppercase{Yiwei Wang}} is an Assistant Professor in the Department of Mathematics at the University of California, Riverside. His research interests include mathematical modeling and scientific computing with applications in physics, material science, biology, and data science. His email address is \email{yiwei.wang@ucr.edu} and his website is \url{https://profiles.ucr.edu/app/home/profile/yiweiw}.\\

\noindent {\bf \MakeUppercase{Chun Liu}} is a Professor and Chair of the Department of Applied Mathematics at the Illinois Institute of Technology, Chicago, IL. His research interests center around partial differential equations, calculus of variations, and their applications in complex fluids. His e-mail address is \email{cliu124@iit.edu} and his website is \url{https://www.iit.edu/directory/people/chun-liu}.

\end{document}